%% file: usopcorr.tex
\documentclass[12pt]{article}
\usepackage{chicagob}
\usepackage{times}

\newcommand{\Lg}{L^g}
\newcommand{\dg}{\delta^g}
\newcommand{\de}{\eta}
\newcommand{\dign}{\delta_{\mathrm{ign}}}
\newcommand{\drules}{{\cD(\cX, \rand)}}
\newcommand{\dgrules}{{\cD(\cX, \rand, \B)}}
\newcommand{\dprules}{{\cD_{\Pr}(\cX, \rand, \B)}}
\newcommand{\der}{{\cal E}}
\newcommand{\derules}{{\der(\cP, \cX, \B)}}
\newcommand{\cPy}{{{\cal P}}}
\newcommand{\Py}{{\Pr_{\cY}}}
\newcommand{\Qy}{{Q}}

\newcommand{\Px}{{P}}
\newcommand{\rand}{{\cal A}}
\newcommand{\dprior}{\mbox{$\delta_{\mbox{\tiny prior}}$}}
\newcommand{\dpost}{\mbox{$\delta_{\mbox{\tiny post}}$}}

\newcommand{\interior}{{\tt int}}
\newcommand{\dcon}{\mbox{$\delta_{\mbox{\tiny con}}$}}

\newcommand{\dgcon}{\mbox{$\delta^{\mbox{\tiny g}}_{\mbox{\tiny con}}$}}
\newcommand{\dglo}{\mbox{$\delta_{\mbox{\tiny glo}}$}}

\newcommand{\argmin}{\mbox{argmin}}

\newcommand{\cP}{{\cal P}}
\newcommand{\hull}[1]{\langle #1 \rangle}
\newcommand{\cU}{{\cal U}}
\newcommand{\reals}{{\IR}}

\newcommand{\cB}{{\cal B}}
\newcommand{\cD}{{\cal D}}
\newcommand{\cX}{{\cal X}}
\newcommand{\cY}{{\cal Y}}

\newenvironment{oldthm}[1]{\par\noindent{\bf Theorem #1:} \em \noindent}{\par}
\newenvironment{oldlem}[1]{\par\noindent{\bf Lemma #1:} \em \noindent}{\par}
\newenvironment{oldcor}[1]{\par\noindent{\bf Corollary #1:} \em \noindent}{\par}
\newenvironment{oldpro}[1]{\par\noindent{\bf Proposition #1:} \em \noindent}{\par}
\newcommand{\othm}[1]{\begin{oldthm}{\ref{#1}}}
\newcommand{\eothm}{\end{oldthm} \medskip}
\newcommand{\olem}[1]{\begin{oldlem}{\ref{#1}}}
\newcommand{\eolem}{\end{oldlem} \medskip}
\newcommand{\ocor}[1]{\begin{oldcor}{\ref{#1}}}
\newcommand{\eocor}{\end{oldcor} \medskip}
\newcommand{\opro}[1]{\begin{oldpro}{\ref{#1}}}
\newcommand{\eopro}{\end{oldpro} \medskip}
\newtheorem{conjecture}{Conjecture}
\input{defn}


\title{A Game-Theoretic Analysis of Updating Sets of Probabilities}

\author{{\bf Peter D. Gr{\"u}nwald}\\
CWI, P.O. Box 94079 \\ 1090 GB Amsterdam\\
pdg@cwi.nl\\
http://www.grunwald.nl
\and
{\bf Joseph Y.\ Halpern}\\
Cornell University\\
Ithaca, NY 14853\\
halpern@cs.cornell.edu\\
http://www.cs.cornell.edu/home/halpern
}
\begin{document}

\date{}
\maketitle

\begin{abstract}
We consider how an agent should update her uncertainty when it is
represented by a set $\P$ of probability distributions and the agent
observes that a random variable $X$ takes on value $x$,  
given that the agent makes decisions using the \emph{minimax criterion},
perhaps 
the best-studied and most commonly-used criterion in the literature.
We adopt a game-theoretic framework, where the agent plays
against a bookie, who chooses some distribution from $\P$. We consider
two reasonable games that differ in what the bookie knows when he makes 
his choice.  
Anomalies that have been observed
before, like \emph{time inconsistency}, can be understood as arising
because different games are being played, against bookies with different
information. We characterize the important
special cases in which the optimal decision rules 
according to the minimax criterion
amount to 
either conditioning or simply ignoring the information. Finally, we
consider the 
relationship between conditioning and \emph{calibration} when
uncertainty is described by sets of probabilities.
\end{abstract}

\commentout{
  Dear Joe, \\
  We declared victory too early! Our main conjecture turned out to be
  false. However, we still have a lot of material, probably enough for
  an interesting paper. So what I did is: I tried to write down
  everything we have this far (including a counterexample to our
  original conjecture), sticking as much as possible to our original
  plans for the write-up. The resulting document is long but contains
  really everything we have, so this time, it would be good if you could read
  all of it. To make it a nice story, we might have to change the
  order of many things, but I think it should be possible.

Our
  original plan was as follows. After introducing the general setting and explaining there is no single optimal way to proceed for an agent, 
we consider in turn what an agent should do in three different situations/under three different critera: fixed loss, adversarial loss, calibration.
In terms of sections:
\begin{enumerate}
\item Introduction of what I call $\phi$-conditioning below.
  (Conditioning on a coarsening of $X$). 
\item Discuss First Criterion: minimax optimal decision relative to fixed loss
  function. We show that the minimax decision rule may be interpreted
  as $\phi$-conditioning for some $\phi$ depending on $\cP$ and $L$,
  and then picking the minimax act relative to the conditional
  distributions obtained. {\bf I discovered that unfortunately this is
    false!}
\item Second Criterion: loss function
  determined by an adversary. We show that the condition-on-$X$ (no
  coarsening) rule is optimal. (This works)
\item Third Criterion: calibration: we show that the condition-on-$\phi(X)$
  update rules are the most `refined' update rules that are guaranteed
  to be calibrated. (This works)
\end{enumerate}
So, Section 2 now gives an elaborate counterexample to our claim,
rather than the original claim (in Subsection 2.2 however, I indicate
that minimax decisions can be related to conditioning in some other
way, which I thought of later). Sections 3 and 4 are as we
planned before, i.e. as described above. I also added something about
semi-adversarial bookies in an extra Section 5, I couldn't resist
(actually I think it makes sense to put it in: it makes clear that the
developments in Section 3 and 4 are much more closely related than it
may seem. But feel free to shorten or even remove it). Section 1 tells
(a) the general story, and then (b) focusses on notation and
definitions.  Regarding (a), since our main idea was wrong, we may
want to change this completely (and then also change the order of the
remainder of the paper). Regarding (b), there is quite a lot of
notation/definition, and notation is not 100 \% consistent yet -- we
can probably greatly simplify all of that at a later stage.
\end{abstract}
}

\section{Introduction}
\label{sec:introduction}

Suppose 
that
an agent models her uncertainty about a domain using a {\em
set\/} $\cP$ of probability distributions.
How should the agent make decisions?
Perhaps the best-studied and most commonly-used approach in the
literature is to use the minimax criterion \cite{Wald50,GS82,GS1989}.
According to the minimax criterion, 
action $a_1$ is preferred to action $a_2$ if the
worst-case expected loss of $a_1$ (with respect to all the probability
distributions in the set $\cP$ under consideration) is better than the
worst-case expected loss of $a_2$.  
Thus, the action chosen is the one with the best worst-case outcome.

We are often interested in making decisions, not just in a static
situation, but in a more dynamic situation, where the agent may make
some observations, or learn some information.
This leads to an obvious question:
If the agent represents her uncertainty using a set $\cP$ of probability
distributions, how should she update $\cP$ in light of observing that
random variable $X$ takes on value $x$?
Perhaps the standard 
answer is to condition each distribution in $\cP$ 
on $X=x$
(more precisely, to condition those distributions in $\cP$ that give
 $X=x$ positive probability on $X=x$),
and adopt
the resulting set of conditional distributions $\cP \mid X =
x$
as her representation of uncertainty.
As has been pointed out by several authors, this
sometimes leads to a phenomenon called {\em dilation\/}
\cite{Augustin03,CozmanWalley,HerronSW97,SeidenfeldW93}: 
the agent may have substantial knowledge about some other random variable $Y$
before observing $X=x$, but
know significantly less after conditioning.  
Walley
\citeyear[p. 299]{Walley91} gives a simple example of dilation: suppose
that a fair 
coin is tossed twice, where the second toss may depend in an arbitrary
way on the first.  (In particular, the tosses might be guaranteed to be
identical, or guaranteed to be different.)  If $X$ represents the
outcome of the first toss and $Y$ represents the outcome of the second
toss, then before observing $X$, the agent believes that the probability
that $Y$ is heads is $1/2$, while after observing $X$, the agent
believes that the probability that $Y$ is heads can be an arbitrary
element of $[0,1]$. 

While, as this example and others provided by Walley show, such dilation
can be quite reasonable, it interacts rather badly with the minimax
criterion, leading to anomalous behavior that has been 
called \emph{time inconsistency} \cite{GrunwaldH04,Seidenfeld04}:
the minimax-optimal conditional
decision rule before the value of $X$ is observed (which has the form
``If $X=0$ then do $a_1$; if $X=1$ then do $a_2$; \ldots'') may be
different from the minimax decision rule obtained after conditioning.  
For example, the minimax-optimal conditional decision rule may say ``If
$X=0$ then do $a_1$'', but the minimax-optimal decision rule conditional
on observing $X=0$ may be $a_2$.  (See Example~\ref{xam:inconsistency}.)
If uncertainty is modeled using a single distribution,
such time inconsistency cannot arise. 
\commentout{
Indeed, in earlier work \cite{GrunwaldH04} (GH from now on), we showed
that there are 
settings where it is not the right thing to do.  In this paper, we
consider the question in more detail, and try to characterize more
generally what the right thing to do is.

Following GH,
we adopt a decision-theoretic
framework where we assume that, after observing $X$, the agent has to
perform some action, typically a prediction concerning the value of
some other random variable $Y$. 
The first issue to consider is how we determine whether one action is
better than another.  As in GH, we 
mainly focus on
the minimax
criterion:

However, as is well known, 
if we combine minimax decision rules with conditioning, we run into a
second problematic phenomenon called {\em time inconsistency\/}
\cite{GrunwaldH04,Seidenfeld04}: the minimax-optimal conditional
decision rule before the value of $X$ is observed (which has the form
``If $X=0$ then do $a_1$; if $X=1$ then do $a_2$; \ldots'') may be
different from the minimax decision rule obtained after conditioning.  
For example, the minimax-optimal conditional decision rule may say ``If
$X=0$ then do $a_1$'', but the minimax-optimal decision rule conditional
on observing $X=0$ may be $a_2$.  (See Example~\ref{xam:inconsistency}.)
If uncertainty is modeled using a single distribution,
such time inconsistency cannot arise. 
}

To understand this phenomenon better,
we model the decision problem as a game between the agent and a
bookie.  
It turns out that there is more than one possible game that can be
considered, depending on what information the bookie has.  
We focus on two (closely related) games here.
In the first game,
the bookie chooses a distribution from $\P$ before the agent moves.  
We show that the Nash equilibrium of this game leads to a minimax
decision rule.  (Indeed, this can be viewed as a justification of using
the minimax criterion).  However, in this game,
conditioning on the information is not always optimal.%
\footnote{In some other senses of the words ``conditioning'' and
  ``optimal,'' conditioning on the information {\em is\/} always
  optimal. This is discussed further in Section~\ref{sec:discussion}.} 
\commentout{
\footnote{Walley \citeyear{Walley91} shows that, in a sense,
conditioning is the only updating rule that is \emph{coherent},
according to his notion of coherence.  He justifies coherence decision
theoretically, but not by using the minimax criterion.  Our focus here is on
the interaction between condition and the minimax criterion.}
}

In the second game, the bookie gets to choose the distribution
\emph{after} the value of $X$ is observed.  
Again, in this game, the Nash equilibrium leads to the use of minimax,
but now conditioning \emph{is} the right thing to do.  

If $\P$ is a singleton, the two games coincide (since there is only one
choice the bookie can make, and the agent knows what it is).  Not
surprisingly, conditioning is the appropriate thing to do in this case.
The moral of this analysis is that, when uncertainty is characterized by
a set of distributions, if the agent is making decision using the
minimax criterion, then the right decision depends on the game being
played.  The agent must consider if she is trying to protect herself
against an adversary who knows the value of $X=x$ when choosing the
distribution or one that does not know the value of $X=x$.

In earlier work \cite{GrunwaldH04} (GH from now on), we essentially
considered the first game, and showed that, in this game, conditioning
was not always the right thing to do when using the minimax criterion.  
Indeed, we showed there are sets $\P$ and games for
which the minimax-optimal decision rule is to simply ignore the information.
Our analysis of the first game lets us 
go beyond GH here in 
two
ways. First, we  characterize
exactly when it is minimax optimal to ignore information.  Second, 
we provide a simple sufficient condition for when conditioning on the
information is minimax optimal. 
\commentout{
We then consider a third game.  As in the first game,
the bookie chooses a distribution from ${\cal P}$ before the
value of $X$ is observed. The difference is that the agent now has to
update the set of distributions ${\cal P}$ without knowing the loss
function in advance. 
The loss function is determined by the bookie at a later stage,
without knowing the realized value of $X$.
This
corresponds to a situation that is often encountered in, for example,
statistics (see Example~\ref{ex:science}). 
In this
case, the right thing to do is to condition, but not necessarily on
all the given information. Whereas standard conditioning on $X=x$
implies a restriction of the set of
possible worlds (outcomes) to those in which $X=x$, in
the third game, the agent may end up restricting the set of possible
worlds to a 
superset of those in which $X=x$. We call this generalization of
conditioning ${\cal C}$-conditioning.  
The set on which to
condition is determined by the set ${\cal P}$. In some cases, it is
equal to the set of worlds in which $X=x$; then ${\cal
  C}$-conditioning amounts to ordinary conditioning. In other cases,
it is equal to the set of all possible worlds; then ${\cal
  C}$-conditioning amounts to ignoring the information.  It may now
seem that, perhaps even for the first game, ${\cal C}$-conditioning is
always the optimal thing to do.  
After all, GH showed that in some cases, the minimax-optimal decision is to
ignore the information, 
while in others it is to condition on $X=x$, both of which are 
instances of ${\cal C}$-conditioning. 
However, 
the well-known Monty Hall Problem  (Example~\ref{xam:montyhall}) shows
that this conjecture is false; 
there are times when
the minimax-optimal decision rule in the first game cannot be
understood in terms of conditioning.
}

Ignoring the information
can be viewed as 
the result of conditioning; not conditioning on the information, but
conditioning on the
whole space.  This leads to a natural question: 
suppose that when we observe $x$, we condition on the event that $X
\in \C(x)$, where  $\C(x)$ is some set containing $x$,
but not necessarily equal to $\{x\}$. Is this variant of
conditioning,
an approach we call \emph{$\C$-conditioning}, always minimax
optimal in the first game?  
That is, is it always optimal to condition on \emph{something}?
As we show by considering the well-known
Monty Hall Problem  (Example~\ref{xam:montyhall}), this is not the case
in general.  
Nevertheless, ${\cal C}$-conditioning has some interesting properties:
it is closely related to the concept of  \emph{calibration}
\cite{Dawid82}. 
Calibration is usually defined in terms of empirical data. To explain
what it means, consider an agent that is a weather forecaster on your
local television station. Every night the forecaster makes a
prediction about whether or not it will rain the next day in the area
where you live. She does this by asserting that the probability of
rain is $p$, where $p \in \{0,0.1, \ldots, 0.9,1 \}$. How should we
interpret these probabilities? The usual interpretation
is that, in the long run, on those days at which the weather
forecaster predict probability $p$, it will rain approximately $100 p
\%$ of the time \cite{Dawid82}. Thus, for example, among all days for
which she predicted $0.1$, the fraction of days with rain was close to
$0.1$.
A weather forecaster with this
property is called {\em calibrated}. 
\commentout{
If a weather forecaster is
calibrated, and you make bets which, based on her probabilistic
predictions, seem favorable, then in the long run you cannot lose
money. If a weather forecaster is not calibrated, there exist bets
that may seem favorable but result in a loss.
This suggests that a good probabilistic prediction should be calibrated.
But calibration is only a {\em minimal\/} requirement.
If in fact it rains 30\% of the days of the year, then a 
weather forecaster who predicts $0.3$ for every single day of the year
is
calibrated, but still not very informative. Thus, given two
calibrated forecasters, we prefer the one that makes more ``informative''
predictions, in a sense we make precise.
}

Up to now, calibration has been considered only when uncertainty is
characterized by a single distribution.  We generalize the notion of
calibration to our setting, where uncertainty is characterized by a set
of distributions.  
We then show that a rule for updating a set of probabilities is
guaranteed to be calibrated if and only if it is an instance of ${\cal
  C}$-conditioning. 
In combination with our earlier results, this implies that if 
calibration is considered essential, then an
update rule may sometimes result in decisions 
that are not minimax optimal.

The rest of this paper is organized as follows.  In
Section~\ref{sec:notation}, we define the basic framework.
In Section~\ref{sec:game1}, we formally define the two games described
above and show that the minimax-optimal decision rule gives a Nash
equilibrium.  
In Section~\ref{sec:char}, we characterize the minimax-optimal
decision rule for the first game, in which the bookie chooses a
distribution before $X$ is observed. 
In Section~\ref{sec:calibration}, we discuss 
$\C$-conditioning and calibration.

\section{Notation and Definitions}\label{sec:notation}

\paragraph{Preliminaries:}

For ease of exposition, we assume throughout this paper that we are
interested in two random variables, $X$ and
$Y$, which can take values in spaces $\cX$ and $\cY$, respectively.  
$\cP$ always denotes a set of distributions on
$\cX \times \cY$;
that is, $\cP \subseteq \Delta(\cX \times \cY)$, where, as usual,
$\Delta(S)$ denotes the set of probability distributions on $S$.
For ease of exposition, we assume that $\P$ is a closed set; this is a
standard assumption in the literature that seems quite natural in our
applications, and makes the statement of our results simpler.
If $\Pr \in \Delta(\cX \times \cY)$, let
$\Pr_{\cX}$ and $\Pr_{\cY}$ denote the marginals of $\Pr$ on $\cX$ and $\cY$,
respectively. 
Let $\cP_{\cY} = \{\Pr_{\cY}: \Pr \in \cP\}$.
If $E \subseteq \cX \times \cY$, then let 
$\cP \mid E = \{\Pr \mid E : \Pr \in \cP, \Pr(E) > 0\}$.
Here $\Pr \mid E$ (denoted by some authors as $\Pr(\cdot
\mid E)$) is the distribution on $\cX \times \cY$
obtained by conditioning on $E$. 
\paragraph{Loss Functions:}
As in GH,
we are interested in an agent who must choose some action
from a set $\cA$, where 
the loss 
of the action depends only on the
value of random variable $Y$.
For ease of exposition, we assume in this paper that 
$\cX$, $\cY$, and $\cA$ are always finite. 
We assume that with each action $a \in \cA$ and value $y \in \cY$ is
associated some loss to the agent.  (The losses can be negative, which
amounts to a gain.)  Let 
$L: \cY \times \cA \rightarrow \IR$ 
be the loss function.\footnote{We could equally well use
  utilities, which can be viewed as a positive measure of gain.
  Losses seem to be somewhat more standard in this literature.} 

Such loss functions arise quite naturally. For example,
in a medical setting, we can take $\cY$ to consist of the
possible diseases and $\cX$ to consist of symptoms.  The set $\cA$
consists of possible courses of treatment that a doctor can choose.
The doctor's loss function depends only on the patient's disease and
the course of treatment, not on the symptoms.  But, in general, the
doctor's choice of treatment depends on the symptoms observed.

\commentout{
For finite $\cA$, we sometimes consider
{\em randomized actions\/} $R \in \Delta(\cA)$, and we abbreviate
$E_{\Py,R} [L(Y,A)] = \sum_{y \in \cY, a \in \cA} \Py(y) R(a) L(y,a)$
to $E_{\Py}[L(Y,R)]$.

We restrict our analysis to loss functions that are mathematically
well-behaved. We call such loss functions{\em simple\/}:  
\dfn 
\label{def:simple}
We call a loss function $L$ {\em simple\/} if
it is bounded from below, i.e. 
$$\min_{y \in \cY} \min_{a \in \cA} L(y,a) >
- \infty,$$ 
and either one of the following is the case:
\begin{enumerate}
\item either $\cA$ is finite and $L$ is bounded
\item or else 
\begin{enumerate}
\item $\cA$ is a compact convex subset of $\IR^m$ for some $m
  > 0$, with nonempty interior, and 
\item for each $y \in \cY$, $L(y,a)$ is a
  convex, continuous function of $a$ on $\cA$ in the standard topology
  on $\IR \cup [\infty]$, with $L(y,a) < \infty$ if $a$ is in the
  interior of $\cA$. 
\end{enumerate}
\end{enumerate}
\edfn
Note that if $\cA$ is infinite, then the loss function must be convex, but not necessarily strictly convex. Condition 2(b) implies that losses may
  only become infinite at the boundary of the action space. 

For finite $\cY$ (which we assume), most loss functions studied in the
literature are simple: losses involved in point predictions and/or
betting games correspond to finite $\cA$, losses considered in the
statistical literature such as the logarithmic score or the squared
loss function (Brier score) correspond to compact $\cA$.  Both of
these are examples of loss functions $L$ that, adjusting terminology
from \cite{GrunwaldD04}) to our situation, we call {\em strictly
  proper}. Formally:

\dfn
  A simple loss function $L$ is {\em strictly proper\/} if (a) $\cA =
  \Delta(\cY)$; (b) for each $\Py \in \Delta(\cY)$, $E_{\Py}[L(Y,a)]$ is
  uniquely minimized by $a = \Py$.
\edfn
\paragraph{Minimax Actions:}
The proof of the following proposition follows by simple continuity
arguments and is omitted. 
\pro
\label{pro:mmactexists}
Let $L$ be simple.
\begin{itemize} 
\item[a.] Let $\Py \in \Delta(\cY)$. Then $\min_{a \in \cA}
  E_{\Py}[L(Y,a)] = \min_{R \in \Delta(\cA)} E_{\Py}[L(Y,R)]$ is finite and
  achieved by a (not necessarily unique) {\em Bayes act\/} $a \in \cA$.
\item[b.] Let $\cPy \subseteq \Delta(\cY)$ be closed. Then $\min_{R \in
    \Delta(\cA)} \max_{{\Py} \in \cPy} E_{\Py}[L(Y,R)]$ is finite and
  achieved by some (not necessarily unique) {\em minimax act\/} $R \in
  \Delta(\cA)$.
\begin{enumerate}
\item If $\cA$ is finite, then the set $\{R^* : E_{\Py}[L(Y,R^*)] =
  \min_{R \in \Delta(\cA)} \max_{{\Py} \in \cPy} E_{\Py}[L(Y,R)]\}$ is
  convex and closed.
\item If $\cA$ is infinite, then $\min_{a \in \cA} \max_{{\Py} \in
    \cPy} E_{\Py}[L(Y,a)] = \min_{R \in \Delta(\cA)} \max_{{\Py} \in
    \cPy} E_{\Py}[L(Y,R)]$ is achieved by some deterministic action
  $a^* \in \cA$, and the set $\{a^* : E_{\Py}[L(Y,a^*)] = \min_{a \in
    \cA} \max_{{\Py} \in \cPy} E_{\Py}[L(Y,a)]\}$ is convex and closed.
\item If $\cA$ is infinite and $L$ is strictly convex, then the
  minimax act $a^*$ is unique.
\end{enumerate}
\end{itemize}
\epro 
In this paper, we are interested in
characterizing actions that are minimax under some conditions on
$\cPy$. The proposition shows that unless $\cA$ is finite, there is never any need for such minimax actions to be randomized. Therefore, it is convenient
to define the `effective' set of actions $\rand$, where
$\rand = \Delta(\cA)$ if $\cA$ is finite, and $\rand = \cA$
otherwise, and to define a minimax act as any $a \in \rand$
achieving $\min_{a \in \rand} \max_{\Py \in \cPy} E_{\Py}
[L(Y,a)]$.  With this convention, 
a loss function $L$ with finite $\cA$ becomes indistinguishable
from a loss function $L'$ defined with respect to the infinite set 
$\cA' = \Delta(\cA)$, with $L'(y,R) := \sum_{a \in \cA} R(a)L(y,a)$. 

It will sometimes be convenient to speak of `the' minimax
act $a^* \in \rand$, even though there may in fact be several
ones. Our results do not depend on the particular $a^*$ we pick, but
for definiteness, let us define, for given $\cPy$,
\begin{equation}
\label{eq:argmin}
a^* := \argmin_{a \in \rand} {\max}_{\Py \in \cPy} E_{\Py}[L(Y,a)]
\end{equation}
as the center of mass of the (convex, nonempty) set of minimax acts
relative to $\cPy$.
} %

\paragraph{Decision Rules:}
Suppose that the agent observes the value of a variable
$X$ that takes on values in $\cX$. After having observed $X$, she must
perform an act, the quality of which is judged according to loss
function $L$. 
The agent must choose a {\em decision rule\/} that
determines what she does as a function of her observations.
We allow decision rules to be randomized.  Thus, a decision rule is a
function $\delta: \cX \rightarrow \Delta({\cal A})$ that chooses a
distribution over actions based on the agent's observations.
Let $\drules$ be the set of all decision rules.
A special case is a
deterministic decision rule, which assigns probability 1 to a particular
action.  If $\delta$ is deterministic, we sometimes abuse notation and
write $\delta(x)$ for the action that is assigned probability 1 by the
distribution $\delta(x)$.
Given a decision rule $\delta$ and a loss 
function $L$, let $L_\delta$ be the random variable on $\cX \times \cY$
such that $L_\delta(x,y) = \sum_{a \in \cA} \delta(x)(a) L(y,a)$.
Here $\delta(x)(a)$ stands for the probability of performing action $a$
according to the distribution $\delta(x)$ over actions that is adopted
when $x$ is observed.
Note that in 
the special case that $\delta$ is a deterministic decision rule,
$L_\delta(x,y) = L(y, \delta(x))$.  

A decision rule 
$\delta^0$ 
is \emph{a priori minimax optimal} with
respect to $\cP$ 
and $\cA$ if 
\begin{equation}
\label{eq:priora}
{\max}_{\Pr \in \P} E_{\Pr}[L_{\delta^0}] = 
{\min}_{\delta \in  \drules} {\max}_{\Pr \in \P} E_{\Pr}[L_\delta].
\end{equation}
That is, $\delta^0$ is a priori minimax optimal if $\delta^0$ gives the best
worst-case expected loss with respect to all the distributions in
$\Pr$.%
\commentout{
\footnote{Note that the minimax criterion puts a total order on decision
rules.  That is, we can say that $\delta$ is at least as good as $\delta'$ if 
$${\max}_{\Pr \in \P} E_{\Pr}[L_\delta] \le {\max}_{\Pr \in \P}
E_{\Pr}[L_{\delta'}].$$
By way of contrast, Walley \citeyear{Walley91} puts a partial order on
decision rules by taking $\delta$ to be at least as good as $\delta'$ if
$${\max}_{\Pr \in \P} E_{\Pr}[L_\delta - L_{\delta'}] \le 0.$$  
Since both ${\max}_{\Pr \in \P} E_{\Pr}[L_\delta - L_{\delta'}]$ and 
${\max}_{\Pr \in \P} E_{\Pr}[L_{\delta'} - L_{\delta}]$ may be
positive, this is indeed a partial order.  If we use this ordering to
determine the optimal decision rule then, as Walley shows, conditioning
is the only right thing to do. }
}
Note that we can write max here instead of sup because of our assumption
that $\P$ is closed.  This ensures that there is some $\Pr \in \P$ for
which $E_{\Pr} [L_{\delta^0}]$ takes on its maximum value.

A decision rule 
$\delta^1$ 
is \emph{a posteriori minimax optimal} with
respect to $\cP$ and $\cA$ if, for all $x \in \cX$
such that $\Pr(X=x) > 0$ for some $\Pr \in \cP$,
\begin{equation}
\label{eq:posteriora}
\begin{array}{l}
{\max}_{\Pr \in \cP \mid X=x}E_{\Pr}[L_{\delta^1}] = \\
{\min}_{\delta \in  \drules} {\max}_{\Pr \in \cP \mid X=x}
E_{\Pr}[L_\delta]. 
\end{array}
\end{equation}
To get the a posteriori minimax-optimal decision rule we do the
obvious thing: if $x$ is observed, we simply condition each
probability distribution $\Pr \in \P$ on $X=x$, and choose the action
that gives the least expected loss (in the worst case) with respect to
$\cP \mid X=x$.%
Note that all distributions $\Pr$ mentioned in (\ref{eq:posteriora})
satisfy $\Pr(X=x) = 1$. Therefore, the minimum over $\delta \in
\drules$ does not depend on the values of $\delta(x')$ for $x' \neq
x$; the minimum is effectively over randomized actions rather than
decision rules.

\commentout{
It will be convenient to introduce mathematical symbols for a priori
and a posteriori minimax decision rules. We define the a priori
minimax decision rule $\dprior$ as
\begin{equation}
\label{eq:priorb}
\dprior = \argmin_{\delta \in  \drules} {\max}_{\Pr \in \P}
E_{\Pr}[L_\delta],
\end{equation}
extending the definition of $\argmin$ (\ref{eq:argmin}) for minimax
acts to minimax decisions in the obvious manner.
We define the a priori minimax rule $\dpost$ as 
\begin{equation}
\dpost(x)
\label{eq:posteriorb}
= \argmin_{a \in  \Delta({\cal A})} {\max}_{\Pr \in \cP \mid X=x}
E_{\Pr}[L(Y,a)].
\end{equation}
} %
As the following example, taken from GH, shows, a
priori minimax-optimal decision rules are in general different from
a posteriori minimax-optimal decision rules.
\xam\label{xam:inconsistency} Suppose that 
$\cX = \cY = \cA = \{0,1\}$ and $\cP = \{\Pr \in \Delta(\cX \times \cY) :
\Pr_{\cY}(Y=1) = 2/3\}$.  
Thus, $\P$ consists of all distributions whose marginal on $Y$ gives
$Y=1$ probability $2/3$.
We can think of the actions in $\cA$ as predictions of the value of $Y$.
The loss function is 0 if the right value is predicted and 1 otherwise;
that is, 
$L(i,j) = |i-j|$. This is the so-called $0/1$ or {\em classification\/} loss.
It is easy to see that the optimal a priori decision rule is to choose 1
no matter what is observed (which has expected loss $1/3$).    
Intuitively, observing the value of $X$ tells us nothing about the value
of $Y$, so the best decision is the one which predicts according to the
prior probability of $Y=1$.
However, all probabilities on $Y=1$ are compatible with observing either
$X=0$ or $X=1$.  That is, both $(\cP \mid X=0)_{\cY}$ and $(\cP \mid 
X=1)_{\cY}$ consist of all distributions on $\cY$.  
Thus, the minimax
optimal a posteriori decision rule randomizes (with equal probability)
between $Y=0$ and $Y=1$.  

Thus, if you make decisions according to the minimax rule, then before
making an observation, you will predict $Y=1$.  However, \emph{no matter
what observation you make}, after making the observation, you will
randomize (with equal probability) between predicting $Y=0$ and $Y=1$.
Moreover, you know even before making the observation that your opinion
as to the best decision rule will change in this way.
\exam
\commentout{

\paragraph{Probability Update Rules:}
Many decision rules are obtained on the basis 
of some \emph{probability update rule}. That is, upon observing that
$X=x$, an agent first updates $\cP$ to a set $\cP_x$
(for example, $\cP_x$ could be $\cP \mid X=x$, but does not have
to be), and then chooses the minimax-optimal act with respect to this
set of distributions.  

To make this precise, define a 
{\em probability update rule\/} to be a function $\Pi: 2^{\Delta(\cX \times
\cY)} \times \cX \rightarrow 2^{\Delta(\cX \times \cY)}$.  Intuitively, 
given a set $\cP$ of probability distributions on $\cX \times \cY$ and
an observation $X=x$, $\Pi(\cP,x)$ is the result of updating the set
$\P$ in
light of the observation.  Let $\delta_{\Pi}$ be the decision rule
defined  by taking
\begin{equation}
\label{eq:pibased}
\delta_{\Pi}(x) = 
\arg
{\min}_{a \in \rand} {\max}_{\Pr \in \Pi(\cP,x)} E_{\Pr}[L(Y,a)]
\end{equation}
Thus, the maximum a posteriori decision rule defined earlier can be
viewed as $\delta_\Pi$, where $\Pi$ is just conditioning, that is,
$\Pi(\cP,x) = \P \mid X=x$.  
We call $\delta_{\Pi}$ the decision rule {\em based on update rule $\Pi$}.

We will be interested in one particular type of update rule.  Let $\C =
\{\X_1, \ldots, \X_k\}$ be a partition of $\X$; that is, $\X_i \ne
\emptyset$ for $i = 1, \ldots, k$; $\X_1 \union
\ldots \X_k = \X$; and $\X_i \inter \X_j = \emptyset$ for $i \ne j$.
If $x \in \X$, let $\C(x)$ be the cell containing $x$; 
that is, the unique
element $\X_i \in \C$ such that $x \in \X_i$.  The
\emph{$\C$-conditioning} update rule is defined by taking $\Pi(\P,x) =
\P \mid \C(x)$ (where we identify a subset $\X'$ of $\X$ with the subset $\X'
\times \Y$ of $\X \times \Y$).
Notice that standard conditioning is a $\C$-conditioning, where we take
$\C$ to consist of all singletons.  Ignoring information is also
$\C$-conditioning, where $\C = \{\X\}$.  
We define a decision rule to be \emph{partition-based} if it has the form 
$\delta_\Pi$, where $\Pi$ is a $\C$-conditioning update rule for some
partition $\C$.
}

\commentout{
[TODO: following paragraph not strictly needed, since definitions will re-occur in text]
Two decision rules based on update rules that we will frequently
encounter are the decision rule based on ordinary conditioning,
denoted by $\dcon$, and the decision rule based on ignoring the
information ($\phi$-conditioning with $\phi(x) \equiv 0$):
\begin{eqnarray}
\dcon(x) & := &  {\min}_{a \in \rand} {\max}_{\Pr \in \cP^{(Y)}(\cdot \mid X = x)} 
E_{\Pr}[L(Y,a)] \\ 
\dign(x) & := &  {\min}_{a \in \rand} {\max}_{\Pr \in \cP^{(Y)}} 
E_{\Pr}[L(Y,a)]
\end{eqnarray}
Note that $\dign(x)$ is constant on $x \in \cX$.
\section{Fixed Loss Function}
\label{sec:fixed}
In the special case with $\cP$ containing precisely the distributions
with a given marginal $\Pr_{\cY)}$ on $\cY$, the minimax-optimal decision
rule turns out to be ignoring the value of $X$ no matter what it is
\cite{GrunwaldH04}.  In all other examples Joe and Peter had
considered so far, the minimax-optimal decision rule sometimes
amounted to first conditioning at some $\phi$-level, and then taking
minimax acts. This led Joe and Peter to formulate the following
conjecture (June 1st):
\begin{conjecture}
  Let $\cP$ be an arbitrary set of distributions, $L$ be an arbitrary
  loss function, and $\delta$ be the minimax decision rule relative to
  $L$ and $\cP$. Then there exists a function $\phi: \cX \rightarrow
  \reals$ such that
\begin{eqnarray}
\lefteqn{  \delta(x)  =} &&  \nonumber \\
& \arg {\min}_{a \in \rand} \ \ \ 
{\max}_{\Pr \in \cP^{(Y)}(\cdot
    \mid \phi(X) = \phi(x) ) } E_{\Pr} [L(Y,a)]. & \nonumber
\end{eqnarray}
Thus, the minimax-optimal decision rule is a $\Pi$-based rule, where the
probability update rule $\Pi$ is equal to $\phi$-conditioning at some
level $\phi$. 
\end{conjecture}
This conjecture is false, as the following example demonstrates. The
example shows that the conjecture is even false if we restrict to
`nicely structured' $\cP$, such as convex $\cP$, or all $\cP$ which
share the same marginal distribution on $Y$ -- so there is not much
hope for saving our conjecture in any form!  \xam
}
\section{Two Game-Theoretic Interpretations of $\cP$}\label{sec:game1}
What does it mean that an agent's uncertainty is characterized by a set
$\cP$ of probability distributions?  
How should we understand $\P$?  We give $\P$ a game-theoretic
interpretation here: namely, an adversary gets to choose a distribution
from the set $\P$.%
\footnote{This interpretation remains meaningful in several practical
  situations where there is no explicit adversary;
see the final paragraph of this section.}
But this does not completely specify the game.  
We must also specify \emph{when} the adversary makes the choice.  We
consider 
two times that the adversary can choose: the first is before
the agents observes the value of $\cX$, and the second is after.  
We formalize this as two different games, where we take the
``adversary'' to be a bookie.

We call the first game the $\P$-game.  It is defined as follows:
\begin{enumerate}
\item The bookie chooses a distribution $\Pr \in \P$.
\item The value $x$ of $X$ is chosen (by nature) according to $\Pr_{\cX}$ and
observed by both bookie and agent.
\item The agent chooses an action $a \in \A$.
\item The value $y$ of $Y$ is chosen according to 
$\Pr \mid X=x$.
\item The agent's loss is $L(y,a)$; the bookie's loss is $-L(y,a)$.
\end{enumerate}

\noindent This is a zero-sum game; the agent's loss is the bookie's
gain.  In this 
game, the agent's strategy is a 
decision rule, that is, a function that gives a distribution over actions
for each observed value of $X$.  The bookie's strategy is a
distribution over distributions in $\P$.

We now consider a second interpretation of $\cP$, characterized by a
different game that gives the bookie more power.  Rather than choosing
the distribution before observing the value of $X$, the bookie gets to
choose the distribution after observing the value.  
We call this the $\cP$-$X$-game.

\begin{enumerate}
\item The value $x$ of $X$ is chosen (by nature) in such a way that
$\Pr(X=x) > 0$ for some $\Pr \in \cP$, and observed by both the bookie
and the agent.
\item The bookie chooses a distribution $\Pr \in \P$ such that $\Pr(X=x)
> 0$.%
\footnote{If we were to consider conditional probability measures, for
which $\Pr(Y=y \mid X=x)$ is defined even if $\Pr(X=x) = 0$, then we
could drop the restriction that $x$ is chosen such that $\Pr(X=x) > 0$
for some $\Pr \in \P$.}
\item The agent chooses an action $a \in \A$.
\item The value $y$ of $Y$ is chosen according to $\Pr\mid X=x$.
\item The agent's loss is $L(y,a)$; the bookie's loss is $-L(y,a)$.
\end{enumerate}

Recall that a pair of strategies
$(S_1,S_2)$ is a Nash equilibrium if neither party can do better by
unilaterally changing strategies.    
If, as in our case, $(S_1, S_2)$ is a Nash equilibrium in a zero-sum
game, it is also known as a ``saddle point''; $S_1$ must be a
minimax strategy, and $S_2$ must be a maximin strategy
\cite{GrunwaldD04}. 
As the following results show,
an agent must be using an a priori minimax-optimal decision rule in 
a Nash equilibrium of the $\cP$-game, and an a posteriori minimax-optimal
decision rule is a Nash equilibrium of the $\cP$-$X$-game.
This can be
viewed as a  justification for using (a priori and a posteriori)
minimax-optimal decision rules.

\commentout{
\pro\label{pro:prior} 
In a Nash equilibrium of the $\P$-game, 
the agent
must be using an a priori minimax-optimal decision rule.
\epro
The proposition is a special case of the following theorem, which
characterizes the Nash equilibria and the a priori minimax-optimal
decision rules in more detail. It is a straightforward extension of
standard game-theoretic results, proven in Appendix~\ref{app:dawidproofs}.
}
\begin{theorem}
\label{thm:dawida}
Fix $\cX$, $\cY$, $\cA$, $L$, and $\cP \subseteq \Delta(\cX \times \cY)$.
\begin{itemize}
\item[(a)] The $\P$-game has a Nash equilibrium  
$(\pi^*,\delta^*)$, where $\pi^*$ is a distribution over $\P$ with
finite support.
\item[(b)] If $(\pi^*,\delta^*)$ is a Nash equilibrium of the
$\P$-game
such that $\pi^*$ has finite support,
then 
\begin{itemize}
\item[(i)] 
for every distribution $\Pr' \in \P$ in the support of $\pi^*$, 
 we have
$$
E_{\Pr'} [L_{\delta^*}] = 
{\max}_{\Pr \in \cP} E_{\Pr} [L_{\delta^*}];
$$ 
\item[(ii)] if $\Pr^* = \sum_{\Pr \in \cP, \pi^*(\Pr) > 0} \pi^*(\Pr)
\Pr$ (i.e., $\Pr^*$ is the convex combination of the distributions in
the support of $\pi^*$, weighted by their probability according to
$\pi^*$), then 
$$
\begin{array}{cl}
&E_{\Pr^*} [L_{\delta^*}]\\
= & 
{\min}_{\delta \in \drules} 
E_{\Pr^*} [L_{\delta}]\\  
= &\max_{\Pr \in \cP} {\min}_{\delta \in \drules} 
E_{\Pr} [L_{\delta}] \\
= &  
{\min}_{\delta \in \drules} 
{\max}_{\Pr \in \cP} E_{\Pr} [L_{\delta}] \\
= &{\max}_{\Pr \in \cP} E_{\Pr} [L_{\delta^*}].
\end{array}
$$ 
\end{itemize}
\end{itemize}
\end{theorem}
Once nature has chosen a value for  $X$ in the $\cP$-$X$-game, we can regard 
steps 2--5 of the $\cP$-$X$-game as a game between the bookie and the
agent, 
where the bookie's strategy is characterized by a distribution in $\cP
\mid X=x$ 
and the agent's is characterized by a distribution over actions.
We 
call this the $\cP$-$x$-game. 
\thm\label{thm:posteriorb} 
Fix $\cX$, $\cY$, $\cA$, $L$, $\cP \subseteq \Delta(\cX \times   \cY)$. 
\begin{itemize}
\item[(a)] The $\P$-$x$-game has a Nash equilibrium
$(\pi^*,\delta^*(x))$, where $\pi^*$ is a distribution over $\cP \mid
X=x$ with finite support.
\item[(b)] If $(\pi^*,\delta^*(x))$ is a Nash equilibrium of the
$\P$-$x$-game
such that $\pi^*$ has finite support, 
then 
\begin{itemize}
\item[(i)]
for all $\Pr'$ in the support of $\pi^*$, we have
$$E_{\Pr'} [L_{\delta^*}] = 
{\max}_{\Pr \in \cP \mid X=x} E_{\Pr} [L_{\delta^*}];$$
\item[(ii)] if $\Pr^* = \sum_{\Pr \in \cP, \pi^*(\Pr) > 0} \pi^*(\Pr)
\Pr$, then 
$$
\begin{array}{cl}
&E_{\Pr^*} [L_{\delta^*}]\\
= & 
{\min}_{\delta \in \drules} 
E_{\Pr^*} [L_{\delta}]  \\

= &\max_{\Pr \in \cP \mid X=x} {\min}_{\delta \in \drules} 
E_{\Pr} [L_{\delta}] \\
= &  
{\min}_{\delta \in \drules} 
{\max}_{\Pr \in \cP \mid X=x} E_{\Pr} [L_{\delta}] \\
= &{\max}_{\Pr \in \cP \mid X=x} E_{\Pr} [L_{\delta^*}].
\end{array}
$$
\end{itemize}
\end{itemize}
\end{theorem}
Since all distributions $\Pr$ in the expression 
${\min}_{\delta \in \drules} {\max}_{\Pr \in \cP \mid X=x} E_{\Pr}
[L_{\delta}]$ in part (b)(ii) are in $\cP \mid X=x$, 
as in (\ref{eq:posteriora}), the minimum is effectively over randomized
actions rather than decision rules.

Theorems~\ref{thm:dawida} and~\ref{thm:posteriorb} can be viewed as
saying that there is no time inconsistency; rather, we must just be
careful about what game is being played.  If the $\P$-game is being
played, the right strategy is the a priori minimax-optimal strategy,
both before and after the value of $X$ is observed; similarly, if the
$\P$-$X$-game is being played, the right strategy is the a posteriori
minimax-optimal strategy, both before and after the value of $X$ is
observed. 
Indeed, thinking in terms of the games explains the apparent time
inconsistency.  While it is true that the agent gains more information
by observing $X=x$, in the $\cP$-$X$ game, so does the bookie.  This
information may be of more use to the bookie than the agent, so, in this
game, the agent can be worse off by being given the opportunity to learn
the value of $X$.

Of course, in most practical situations, agents (robots,
statisticians,\ldots) are not really confronted with a bookie
who tries to make them suffer. Rather, the agents may have no idea at
all what distribution holds, except that it is in some set $\P$. Because
they have no idea at all, they decide to prepare themselves for the
worst-case and play the minimax strategy.  The fact that such a minimax
strategy can be interpreted in terms of a Nash equilibrium of a game
helps to understand differences between different
forms of minimax (such as a priori and a posteriori minimax).
{F}rom this point of view, it seems strange to have a bookie 
choose between different distributions in $\P$ according to some
distribution $\pi^*$.  However, if $\P$ is convex, we can replace the
distribution $\pi^*$ on $\P$ by a single distribution in $\P$, which
consists of the convex combination of the distributions in the support
of $\pi^*$; this is just the distribution $\Pr^*$ of
Theorems~\ref{thm:dawida} and~\ref{thm:posteriorb}.  
Thus, Theorems~\ref{thm:dawida} and~\ref{thm:posteriorb}
hold with the bookie restricted to a deterministic strategy.

\section{Characterizing A Priori Minimax-Optimal Decision Rules}
\label{sec:char}
To get the a posteriori minimax-optimal decision rule 
we do
the obvious
thing: if $x$ is observed, we simply condition each probability
distribution $\Pr \in \P$ on $X=x$, and choose the action that gives the
least expected loss  (in the worst case) with respect to $\cP \mid X=x$.

We might expect that the a priori minimax-optimal decision rule should
do the same thing.  That is, it should be the decision rule that
says, if $x$ is observed, then we choose the action that again gives the
best result (in the worst case) with respect to $\cP \mid X=x$.
However, as shown in GH, this intuition is incorrect in general.  There
are times, for example, that the best thing to do is to ignore the
observed value of $X$, and just choose the action that gives the least
expected loss (in the worst case) with respect to $\cP$, no matter what
value $X$ has.  
In this section we first give
a sufficient condition for conditioning to be optimal, and then
characterize when ignoring the observed value is optimal.
\dfn Let 
$\hull{\cP} = \{ \Pr 
\in \Delta(\cX \times \cY): \Pr_{\cX} \in \cP_{\cX}
\mbox{ and } (\Pr \mid X=x) \in (\cP \mid X=x)$ \mbox{ for all $x \in
\cX$ such that $\cP \mid X=x$ is nonempty}\}.
\edfn
Thus, $\hull{\cP}$ consists of all distributions $\Pr$ whose marginal on
$\cX$ is the marginal on $\cX$ of some distribution in $\cP$ and whose
conditional on observing $X = x$ is the conditional of some distribution
in $\cP$, for all $x \in \X$.  
Clearly $\cP \subseteq \hull{\cP}$, but the converse is not
necessarily true. 
When it is true, conditioning is optimal.
\pro 
\label{pro:conditionmm}
If $\cP = \hull{\cP}$, then there exists an a priori
minimax-optimal rule that is also a posteriori minimax optimal. 
If, for all $\Pr \in \cP$ and all $x \in \cX$, $\Pr(X=x) > 0$, then every a
priori  minimax-optimal rule is also a posteriori minimax optimal.\epro
As we saw in Example~\ref{xam:inconsistency}, the minimax-optimal
a priori decision rule is not always the same as the minimax-optimal a
posteriori decision rule.  In fact, the minimax-optimal a priori
decision rule ignores the information observed.  
Formally, a rule $\delta$ \emph{ignores information} if
$\delta(x) = \delta(x')$ for all $x, x' \in \X$.
If $\delta$ ignores information, define $L'_\delta$ to be the random
variable on $\cY$ such that $L'_\delta(y) = L_\delta(x,y)$ for some
choice of $x$.  This is well defined, since $L_\delta(x,y) =
L_\delta(x',y)$ for all $x, x' \in \cX$.

\commentout{
\thm
Let $\cX$ and $\cY$ be finite, $L$ be simple and 
$\cP \subseteq \Delta(\cX \times \cY)$ be closed and convex. Then 
\begin{itemize}
\item[a.] If 
$\max_{\Pr \in \cP} H^X_L(\Pr) = \max_{\Py \in {\cP}_{\cY}} H_L(P)$, then
$\dign$ is a priori minimax optimal. 
\item[b.] Conversely, if $\max_{\Pr \in \cP} H^X_L(\Pr) \neq 
\max_{\Py \in {\cP}_{\cY}} H_L(P)$, then \\
$\max_{\Pr \in \cP} H^X_L(\Pr) < 
\max_{\Py \in {\cP}_{\cY}} H_L(P)$ and $\dign$ is not a priori minimax optimal.
\elem

The condition of this lemma is not always of  direct use since the
quantity $H^X_L(\Pr)$ may be difficult to compute. Indeed, the main use
of the lemma is in establishing the following result, which gives a
necessary and sufficient condition that holds independently of $L$:
}

\thm
\label{thm:ignoremm}
Fix $\cX$, $\cY$, $L$, $\cA$, and $\cP \subseteq \Delta(\cX \times \cY)$.
If,
for all $\Py \in \cP_{\cY}$, $\cP$ contains a distribution
$\Pr'$ such that $X$ and $Y$ are independent under $\Pr'$, and $\Pr'_{\cY} =
\Py$,  
then there is an a priori minimax-optimal decision rule that ignores
information.
Under these conditions,
if $\delta$ is an a priori minimax-optimal decision rule that
ignores information, then $\delta$ essentially optimizes with respect to
the marginal on $Y$; that is, $\max_{\Pr \in \cP} 
E_{\Pr}[L_\delta] = \max_{\Pr_{\cY} \in \cP_{\cY}} E_{\Pr_{\cY}}[L'_{\delta}]$.
\ethm
GH focused on the case that $\cP_{\cY}$ is a singleton 
(i.e., the marginal probability on $Y$ is the same for all distributions
in $\cP$) and for all $x$, $\cP_{\cY} \subseteq
(\cP \mid X = x)_{\cY}$.  It is immediate from Theorem~\ref{thm:ignoremm}
that ignoring information is a priori minimax optimal in this case.
\commentout{
We conclude this section by completely characterizing
a priori minimax decision rules  in terms of an extension of 
Gr{\"u}nwald and Dawid's \citeyear{GrunwaldD04} notion of 
\emph{generalized entropy}.  In Gr{\"u}nwald and Dawid's setting, there
is no set $X$, so they focus on actions (which do not depend on $X$)
rather than decision rules.  We extend their definition to our setting.
\commentout{
\begin{equation}
\label{eq:dawid}
H_L(\Py) := {\inf}_{a \in \cA} E_{\Py}[L(Y,a)] = {\inf}_{a \in \rand} 
E_{\Py}[L(Y,a)],
\end{equation}
where the latter equality follows from Corollary 3.1 of
\cite{GrunwaldD04}.  This generalized entropy is defined for
essentially arbitrary loss functions $L$. The entropy corresponding to
the logarithmic loss (score), is the ordinary Shannon
entropy. For other loss functions the generalized entropy will still
be concave and inherit a number of other pleasant properties of the
Shannon entropy.  The entropy corresponding to the loss functions
considered in this text, e.g.  the one of Example~\ref{xam:oops}, are
piecewise linear concave functions.

Neither the generalized entropy nor the theorems in \cite{GrunwaldD04}
directly apply to our setting: while \citeN{GrunwaldD04} consider a
wide variety of loss functions, sample spaces and action spaces, they
restrict attention to {\em actions\/} (not depending on $X$) rather
than decision rules $\delta: \cX \rightarrow \rand$.  Here we adjust
these ideas to our setting. We redefine the generalized entropy of a distribution $\Pr$ on $\cX \times \cY$,
relative to loss function $L$, as
\begin{equation}
\label{eq:genent}
}
\dfn
Given $\cX$, $\cY$, $L$, and $A$, the \emph{generalized entropy} of loss
function $L$ is a functional on $\delta(\cX \times \cY)$ defined,
denoted
$H^{\cX}_L$, is defined as follows:
$$H^{X}_L(\Pr) := {{\inf}}_{{\delta} \in \drules} E_{\Pr}[L_\delta].$$
\edfn

The Gr{\"u}nwald-Dawid definition of their notion of generalized entropy,
which they denote $H_L$, can be obtained by taking the inf over decision
rules that ignore the information (i.e., those that are constant
functions).  Note that for a lograthmic loss function $L$, $H_L$ is just
that standard Shannon entropy. 

We can now obtain a characterization of a priori minimax-optimal
decision rules. 
\thm
\label{thm:dawidb}
Fix $\cX$, $\cY$, $L$, $\cA$, and $\cP \subseteq \Delta(\cX \times
\cY)$.
Then 
$(\Pr^*,\delta^*)$ is a Nash equilibrium of the $\cP$-game
iff $\Pr^*$ is a generalized maximum entropy distribution.
\ethm
}
\commentout{
The proof is in Appendix~\ref{app:dawidproofs}.  To interpret this
theorem, note that the maximum generalized entropy $\Pr^*$ is the
bookie's analogue of the a priori minimax-optimal decision rule 
$\dprior$. The theorem may be used as a rationale for the maximum
entropy principle and some of its extensions, in some situations
\cite{GrunwaldD04}, but here we focus on two other applications:
\begin{enumerate}
\item The theorem gives a first connection between a priori minimax
  decision rules and conditioning: The Bayes decision rule $\delta^*$
  for $\Pr^*$ amounts to, upon observing $X=x$, choosing the action
  that has minimum expected loss with respect to the conditional
  distribution $\Pr^* \mid X=x$. This means that, if $\cP$ is convex,
  then the a priori minimax decision rule $\dprior = \delta^*$ can be
  interpreted as being based on a form of conditioning, but with
  respect to a single, `maximum entropy' distribution $\Pr^* \in \cP$,
  rather than the entire set $\cP$.
\item Because, irrespective of the loss function of interest,
  generalized entropies must be concave, they are useful tools in
  determining further characterizations of the a priori minimax
  decision rules.  Indeed, Lemma~\ref{lem:ignoremm} and Theorem~\ref{thm:ignoremm}  in Section~\ref{sec:aprioriconditioning}  rely on
  Theorem~\ref{thm:dawidb}.
\end{enumerate}
\subsection{$\C$-conditioning}
\label{sec:aprioriconditioning}
In the previous subsection we showed that the a priori minimax-optimal
decision rule $\dprior$ can be obtained by conditioning the maximum
entropy $\Pr^*$. 
We now investigate whether we can also obtain $\dprior$ by directly
conditioning on the set of distributions $\cP$. Clearly, if $\cP$ is a
singleton distribution, then $\dprior$ amounts to ordinary
conditioning. It follows that there exist sets $\cP$ for which ordinary
conditioning is a priori minimax-optimal. We partially characterize
these $\cP$ in Proposition~\ref{pro:conditionmm}. On the other hand,
Example~\ref{xam:inconsistency} shows that in some cases, standard
conditioning will not work.  In the case studied there, $\dprior$
amounts to ignoring the value of $X$.  Thus, it follows there exist sets $\cP$
for which ignoring $X$ is a priori minimax optimal. We completely
characterize such $\cP$ in Theorem~\ref{thm:ignoremm}.  Since both
standard conditioning and ignoring are forms of $\C$-conditioning,
and also in light of the results of Section~\ref{sec:calibration}, one
may conjecture that $\dprior$ is always a result of $\C$-conditioning
for some $\C$. Example~\ref{xam:oops} at the end of this subsection
shows that this is not the case.
}
\commentout{
\section{Two More Game-Theoretic Interpretations of ${\cal P}$}
\label{sec:game3}
In this section we consider the situation where the agent has to
update her set of distributions ${\cal P}$ without knowing exactly
what the loss function is. 
\xam
\label{ex:science}
This situation occurs frequently in statistics, machine learning,
and modeling in applied sciences such as psychology or econometrics.
For example, suppose that a researcher performs an experiment in order
to find out the distribution of some variable $V$ among some given
population. She observes a large sample of outcomes $v_1, \ldots,
v_n$.  Based on this data, she decides that some probability
distribution $\Pr^*(V = \cdot)$ is a good model for the data. She then
publishes her study and, in particular, $\Pr^*$ in a scientific
journal. The journal is read by some policy maker, who is interested
in predicting the value of $V$ in particular circumstances. In
general, the researcher who publishes $\Pr^*$ is neither aware of the
circumstances in which the policy maker wants to apply $\Pr^*$, nor is
she aware of the exact loss that is incurred when bad predictions of
$V$ are made.  Thus, she cannot recommend specific actions in her
journal publication; she can only summarize her knowledge about the
situation by publishing $\Pr^*$.  \exam This example falls in a
general class of situations where an ``expert'' announces a
probability distribution on a set $\cY$ of potential outcomes. Second,
one or (many) more ``clients'' make actual decisions about $\cY$ based
on $\cP$. Each of these clients may be interested in a different loss
function. The expert may not know what these loss functions are, and
the clients may not know what data $x \in {\cal X}$ the expert based his
predictions on. Other examples include statistical consultancy and weather
forecasting. \xam
\label{ex:weather}
Consider a weather forecaster on your
local television station. Every night the forecaster makes a
prediction about whether or not it will rain the next day in the area
where you live. She does this by asserting that the probability of
rain is $p$, where $p \in \{0,0.1, \ldots, 0.9,1 \}$. The predictions
$p$ may be used by various TV viewers in various situations. Some may
base their decisions  whether or not to go to the beach on the weather
forecaster's predictions. Others may decide whether or not to water a
large area of plants. Translated into our framework, the weather
forecaster observes $X=x$, where $x$ may represent a long sequence of
past measurements of air pressure, temperature, humidity etc. in a
large area. $\cY = \{0,1\}$ represents whether or not it rains. A
prediction such as $p =0.5$ should really be interpreted as a set of
probability distributions $\cP_x = \{ \Pr_{\cY} : 0.45 < 
\Pr_{\cY}(Y=1) \leq 0.55 \}$. Each TV viewer bases her
predictions on her personal loss function $L(\cdot, \cdot, b)$,
where $b$ may be different for different viewers. 
\exam
We model such situations with unknown loss functions in two ways: the {\em generalized $\cP$-game}, which we discuss below, and the {\em restricted generalized $\cP$-game}, which we discuss in Section~\ref{sec:restricted}.
\subsection{The Generalized $\cP$-Game}
The assumption that the loss function is unknown to the agent at the
time of observing $x$ means that it must be determined at a later
stage. In our minimax framework, it is natural to assume that it is
determined by some adversary, one who might have more knowledge than
the agent about the domain under consideration.  We can model this by
adding an extra argument to the loss function $L$: we define a {\em
  generalized loss function\/} as a function $\Lg: \cY \times \cA
\times \cB \rightarrow \reals$, where, intuitively, $\{ \Lg(\cdot,
\cdot, b) : b \in \cB \}$ is the set of (standard) loss functions that
the adversary can choose from. 

The resulting framework may be viewed as a two-player zero-sum game
between an agent (whose moves are actions) and an adversary, who we
again take to be a bookie (whose moves are distributions, that is,
elements of $\cP$, and loss functions, that is,  elements of $\cB$).  We
also assume that, upon observing $X=x$, the agent updates his set of
distributions $\cP$ to some set $\cP_x := \Pi(\cP,x)$, where $\Pi$ is
a {\em probability update rule}.  This is a function $\Pi:
2^{\Delta(\cX \times \cY)} \times \cX \rightarrow 2^{\Delta(\cX \times
  \cY)}$.  Intuitively, given a set $\cP$ of probability distributions
on $\cX \times \cY$ and an observation $X=x$, $\Pi(\cP,x)$ is the
result of updating the set $\P$ in light of the observation.  We
assume that the sets $\cX,\cY,\cA,\cB, \cP$, the generalized loss
function $\Lg(\cdot, \cdot,\cdot)$ and the update function $\Pi$ are
known to both the agent and the bookie.  Below we consider the {\em
  generalized ${\cal P}$-game}, in which we assume that the bookie's
move may depend on the value of $X$.  In Section~\ref{sec:restricted}
we consider a variation of this game that is arguably more realistic, and
in which the bookie does not have such freedom.  For now, we do allow
$b$ to depend on $x$, and the play follows the following protocol:
\begin{enumerate}
\item The bookie chooses some $\Pr \in \cP$ and some $b \in \cB$. 
\item $x$ is generated according to $\Pr_{\cX}$. Its value is observed by
  agent, and  by bookie. 
\item The agent updates the set of distributions $\cP$ to some set $\cP_x := \Pi(\cP, x)$. 
\item Bookie announces $b$ (but not $\Pr$) to the agent. 
\item The agent chooses the randomized action $\alpha \in \Delta(\cA)$
that minimizes her worst-case expected loss under $\cP_x$, i.e. the
action minimizing  
$\max_{\Pr \in \cP_x} E_{\Pr}[\Lg(Y,\alpha,b)]$.
\item The value $y$ of $Y$ is generated according to $\Pr \mid X=x$.
\item The agent's loss is $\Lg(y,a,b)$; the bookie's loss is $-\Lg(y,a,b)$.
\end{enumerate}
In step 5, the agent performs the action that minimizes her expected
loss under the updated set of distributions $\cP_x$.  Thus, we require
the agent's decision-making to consist of two separate steps: first
(before $b$ is known), $\P$ is updated to $\P_x$. Second, when $b$ is
announced, the minimax-optimal decision relative to $\cP_x$ and
$L(\cdot, \cdot,b)$ is taken. This is in accord with the setting
of Example~\ref{ex:science} and~\ref{ex:weather}, where there are really
two agents: an expert (who updates $\cP$ to $\cP_x$ and announces
$\cP_x$, but does not observe $b$) and one or more clients (who are
told $\cP_x$ and know $b$ but not $x$). Since both the expert and
the client want to make good predictions, it is not unreasonable
to view them as a single entity, incurring final loss $\Lg(y,a,b)$.
After all, the researcher's reputation may be affected by the quality
of the outcome.  While this two-stage game is a reasonable model of many
real-world decision processes, one may of course also consider a
variation of this game in which the agent directly maps $x$, $\cP$ and
$b$ to a randomized action, without necessarily first updating $\P$
and then taking a minimax action. This gives the agent more freedom,
and we do not know whether Theorem~\ref{thm:postdom}
and~\ref{thm:semiadv} below continue to hold in this less restrictive
setting.

Theorem~\ref{thm:postdom}  shows that in the generalized
$\P$-game, if the set of loss functions available to the bookie is
sufficiently large, then the agent should always condition on all the
information she has -- if she does not, then the bookie can offer her
bets which she is willing to accept, but which will make her loose
money on average (this is a sort of weakened `Dutch book' type of
argument). In this sense, the generalized $\P$-game has the same
solution as the $\P$-$X$-game of Section~\ref{sec:game1}.  In the
${\cal P}$-$X$-game, the bookie may choose $\Pr \mid X=x$ after $x$
has been chosen, so that it is not very surprising that the agent
should condition on $X=x$. In the
generalized ${\cal P}$-game, the bookie does not have this freedom,
and the result seems less straightforward. To formally state the theorem, we
first need to precisely define the agent and bookie's strategies, as
well as the
conditions on $\Lg$ that are needed to make the theorem hold.
\paragraph{Agent's Strategy:}
In a generalized $\P$-game, an agent's strategy is a {\em generalized
  (agent) decision rule}, which is a function $\dg: \cX \times \cB \rightarrow
\Delta(\rand)$.  We let $\dgrules$ be the set of all generalized agent
decision rules.  Given a generalized agent decision rule $\dg$, a
generalized loss function $\Lg$, and some $b \in \cB$, we define 
 $\Lg_{\dg,b}$ as the random variable on
$\cX \times \cY$ such that $\Lg_{\dg,b}(x,y) = \sum_{a \in \cA}
\dg(x,b)(a) \Lg(y,a,b)$.

We say that a generalized decision rule
$\dg_0$ is {\em based on update rule $\Pi$\/} if for all $\cP \subseteq
\Delta(\cX \times \cY)$, for all $x \in \cX$, all $b \in \cB$, 
\begin{equation}
\label{eq:pibased}
\max_{\Pr \in \Pi(\cP,x)} E_{\Pr}[\Lg_{\dg_0,b}]
= \min_{\dg \in \dgrules} \max_{\Pr \in \Pi(\cP,x)}
E_{\Pr}[\Lg_{\dg,b}].
\end{equation}
For example, $\Pi$ could stand for ignoring $X$, and then $\Pi(\cP,x)
= \P$; or $\Pi$ could stand for ordinary conditioning, and then
$\Pi(\cP,x) = (\P \mid X=x)$. Extending the definition (\ref{eq:posteriora}) of a posteriori
optimality for standard decision rules, we call a generalized decision
rule {\em a posteriori minimax optimal\/} if it is based on ordinary
conditioning, that is, if (\ref{eq:pibased}) holds for $\Pi(\cP,x) = (\P
\mid X=x)$.

We let $\dprules \subseteq \dgrules$ stand for
the set of generalized decision rules that are based on some
probability update rule $\Pi$. It is easily seen that $\dprules \neq
\dgrules$; the inclusion is strict. The agent's strategy in the
generalized $\cP$-game is therefore a generalized decision rule $\dg$, where
$\dg$ is required to lie in the set $\dprules$ of decision rules based
on probability updating.

\paragraph{Bookie's Strategy:}
Analogously to agent, the bookie will be allowed to
randomize over his choice of loss function.  After having observed
$x$, the bookie's strategy thus becomes a {\em bookie decision rule\/}
which maps the chosen distribution $\Pr$ and the observed $x$ to a
distribution on ${\cal B}$. It is thus a function $\de: \cP \times \cX
\rightarrow \Delta(\cB)$.  We let $\derules$ be the set of all bookie
decision rules. Since, before observing $x$, the bookie also has to
choose some $\Pr \in \cP$, the bookie's strategy now becomes a pair
$(\Pr, \eta)$, with $\Pr \in \cP$ and $\eta \in \derules$.

Given a loss function $\Lg$, an agent's decision rule $\dg$, a
bookie decision rule $\de$, and some fixed $\Pr \in \cP$, we let
$\Lg_{\dg,\de}$ be the random variable on $\cX \times \cY$ such that
$$\Lg_{\dg,\de}(x,y) =
\sum_{b \in \cB} \de(\Pr,x)(b) \sum_{a \in \cA} \dg(x,b)(a)
\Lg(y,a,b). 
$$
Then $E_{\Pr}[\Lg_{\dg,\de}]$ denotes the agent's
expected loss if the agent's strategy is $\dg$ and the bookie's
strategy is $(\Pr,\de)$.

\paragraph{Completeness condition on $\Lg$:}
If $\cB$ contains only a single element, then the generalized
$\cP$-game is equivalent to the $\cP$-game. On the other hand, if the
set $\{ L(\cdot, \cdot, b) \mid b \in \cB\}$ contains {\em all\/}
functions from $\cY \times \cA$ to $\reals$, then the agent's minimax
expected loss will be infinite, no matter what strategy she follows.
To avoid these extreme cases, we restrict ourselves to {\em
  complete\/} sets of loss functions.  Formally:
\dfn
\label{def:complete}
Let $\Lg: \cY \times \cA \times \cB \rightarrow \reals$ be a
generalized loss function and let ${\cal L} = \{ \Lg(\cdot,\cdot, b)
\mid b \in \cB \}$ be the corresponding set of standard loss
functions.  Without loss of generality, let $\cA = \{1, \ldots, m\}$.
We say that $\Lg$ is {\em complete\/} if $m > 1$, and there exist
$\gamma_1, \gamma_2 \in \reals$ with $\gamma_1 \leq \gamma_2$ such
that
\begin{description}
\item[C1] For all $L \in {\cal L}$, there exists a $\delta \in \drules$
  such that, (a), for all $x \in \cX$, all $y \in \cY$, 
$L(y,\delta(x)) \leq \gamma_2$, and, (b),
$\max_{\Pr \in \P} E_{\Pr}[L_{\delta}] \leq \gamma_1$.
\item[C2] Let $L \in {\cal L}$, let $r_1, \ldots, r_m \in \reals^m$ and
  $s_1, \ldots, s_m \in \reals^m$ such that for $1 \leq i \leq m$,
  $r_i \geq 0$. Define $L': \cY \times \cA \rightarrow \reals$ by
  setting, for all $a \in \cA$ and $y \in \cY$, $L'(y,a) = r_a L(y,a)
  + s_a$. We require that, if any such $L'$ satisfies the condition {\bf C1}
  above, then $L \in {\cal L}$.
\end{description}
\edfn
The first condition of completeness ensures that the bookie cannot
choose loss functions such that, no matter what the agent does, her
loss becomes either (a) arbitrarily large with high probability, or
(b) arbitrarily large in expectation.  The second condition expresses
that for $\Lg$ to be complete, each loss function $L(\cdot, \cdot, b)$
must correspond to a set of gambles (bets, lottery tickets) on the
outcomes of $Y$ that are offered to the agent by a bookie, where the
bookie has complete freedom to set the price and the amount of the
tickets for sale, as long as there remains a strategy for agent with
bounded loss as required by the first condition.  Complete
sets of loss functions may be very large. For example, for all $\gamma
\in \reals$, the generalized loss function $\Lg$ corresponding to the
set of {\em all\/} loss functions $L$ such that $\max_{a \in \cA,y \in
  \cY} L(y,a) \leq \gamma$ is complete.
Example~\ref{xam:inconsistencyb} below gives another concrete example.

Since we assume $\cA$ and $\cY$ to be finite, we can think of each
$\Lg(\cdot, \cdot, b)$ as a vector in $\reals^{|\cA| \cdot | \cY|}$.
We say that $\{\Lg(\cdot, \cdot, b) \mid b \in \cB \}$ is convex if
the corresponding set of vectors is convex.

We now have all ingredients to state the main theorem of this section,
which says that, if the bookie has the freedom to set the prices, then
a strategy is minimax optimal for agent iff it is an a posteriori
minimax-optimal decision rule. This means that it can be 
obtained by conditioning the set of distributions ${\cal P}$ on $X=x$. 
\begin{theorem} 
\label{thm:postdom}
Fix $\cA, \cX, \cY, \cB, \cP \subseteq \Delta(\cX \times
\cY)$, and some complete $\Lg$. 
A generalized decision rule $(\dg)^*$ is 
a posteriori minimax optimal iff it is a minimax-optimal strategy in the
generalized ${\cal P}$-game, i.e. 
$$
\min_{\dg \in \dprules} 
\max_{\Pr \in \cP} \max_{\eta \in \derules} 
E_{\Pr}[\Lg_{\dg,\eta}]  =
\max_{\Pr \in \cP} \max_{\eta \in \derules} 
E_{\Pr}[\Lg_{(\dg)^*,\eta}].
$$
If $\cP$ is closed and convex and $\{L(\cdot, \cdot, b) \mid b \in
\cB \}$ is convex, then the generalized $\cP$-game has a Nash
equilibrium $((\Pr^*,\eta^*), (\dg)^*)$, i.e.,
$$
\begin{array}{ccl}
E_{\Pr^*}[\Lg_{(\dg)^*,\eta^*}] & = &  
\min_{\dg \in \dprules} E_{\Pr^*}[\Lg_{\dg,\eta^*}]
\\ & = &
\max_{\Pr \in \cP} \max_{\eta \in \derules} 
\min_{\dg \in \dprules} E_{\Pr}[\Lg_{\dg,\eta}] \\ &=& 
\min_{\dg \in \dprules} 
\max_{\Pr \in \cP} \max_{\eta \in \derules} 
E_{\Pr}[\Lg_{\dg,\eta}] \\ &=&
\max_{\Pr \in \cP} \max_{\eta \in \derules} 
E_{\Pr}[\Lg_{(\dg)^*,\eta}].
\end{array}
$$
\end{theorem}
As the following example illustrates, the value of the game,
$E_{\Pr^*}[\Lg_{(\dg)^*,\eta^*}]$, may in general be larger than the
smallest $\gamma_1$ for which $\Lg$ satisfies the completeness condition
{\bf C1}.  
\xam{\bf [Example~\ref{xam:inconsistency}, Cont.]} 
\label{xam:inconsistencyb}
Consider a generalized $\cP$-game with $\cA = \cX = \cY = \cB =
\{0,1\}, \cP= \{ \Pr \mid \Pr_{\cY}(Y=1) = 2/3 \}$, and let $\Lg(i,j,0) =
|i-j|$ and $\Lg(i,j,1) = 1- |i-j|$.  If the bookie were to play $B=0$
no matter what value of $x$ is observed, then this becomes just the
$\cP$-game of Example~\ref{xam:inconsistency}, in which the agent's
minimax-optimal decision rule is to play $1$ whatever the value of
$x$, and thus, to ignore $x$. This rule achieves expected loss $1/3$,
irrespective of the $\Pr \in\cP$ that was chosen by the bookie. If the
bookie were to play $1$ no matter what value of $x$ is observed, then
the agent's optimal strategy is to play $0$ whatever the value of $x$,
and thus, to ignore $x$ again. Again, this rule achieves expected loss
$1/3$. In the generalized $\cP$-game, where the bookie can choose $b$
as a function of $x$, the agent's minimax-optimal strategy $\delta^*$
is to randomize, and to choose between $a = 0$ and $a=1$ with
probability 1/2, no matter what value of $x$ and $b$ she observes.
This coincides with the minimax a posteriori generalized decision
rule, which is obtained by conditioning on $X=x$.  Note that this rule
achieves expected loss $1/2$, which is larger than the agent's minimax
loss in each of the two $\cP$-game with loss $L(\cdot, \cdot, 0)$ and
loss $L(\cdot, \cdot, 1)$.

One way of making $\Lg$ complete is to add every loss function
$\Lg(\cdot, \cdot, b)$ for which there exists a decision rule $\delta$
with minimax expected loss $\leq 1/3$ (corresponding to $\gamma_1$ in
Definition~\ref{def:complete}) and maximum loss $\leq 1$
(corresponding to $\gamma_2$ in Definition~\ref{def:complete}). Let
$\cB'$ be the resulting set of bookie moves.  We can now apply
Theorem~\ref{thm:postdom}, which implies that conditioning is still a
minimax-optimal strategy for agent. Let us roughly sketch why this is
the case. Note that, in our example, for both $x = 0$ and $x=1$, $(\Pi
\mid X=x)_{\cY} = \P_{\cY}$, where $\P_{\cY} = \Delta(\Y)$.  Now consider a
decision rule $\dg_1$ based on a probability update rule $\Pi_1$ such
that $\Pi_1$ is {\em not\/} standard conditioning. It is possible
that, although $\Pi_1$ is not conditioning, $\dg_1$ is still
equivalent to a decision rule based on conditioning. In that case,
there is nothing to prove. If $\dg_1$ is not equivalent to a rule
based on conditioning, then one can show that, for some $x_0 \in \cX$,
$(\Pi_1(x_0))_{\cY}$ excludes an area around the boundary of
$\Delta(\cY)$, i.e. there exists some $\epsilon > 0$ such that either
(a) the set $\{ \Pr_{\cY} : \Pr_{\cY}(Y=1) \leq \epsilon \}$ or (b) the set
$\{ \Pr_{\cY} : \Pr_{\cY}(Y=1) \geq 1- \epsilon \}$ is not included in
$(\Pi_1(x_0))_{\cY}$. We only consider case (a); case (b) is analogous. By
construction of $\Lg$, for each $k > 0$ there exists a $b_k \in \B'$
such that (1) $\Lg(0,0,b_k) = \Lg(1,0,b_k) = 0$; (2) $\max_{\Pr \in
  (\Pi_1(x_0))_{\cY}} E_{\Pr}[\Lg(Y,1,b_k)] < 0$, and (3), for the $\Pr
\in \P_{\cY}$ with $\Pr(Y=1) = 0$, we have $E_{\Pr}[\Lg(Y,1,b_k)] = k$. If
agent uses decision rule $\dg_1$, then bookie can make agent's
expected loss as large as he likes, by choosing a $\Pr \in \cP$ with
$\Pr(Y=1) = 0$, and, upon observing $X=x_0$, offering the loss
function $b_k$ for large enough $k$. For each $k$, the agent will be
induced to choose action $1$, since it seems to give better expected
payoff than action $0$. But in reality, action 1 gives much worse
expected payoff than action $0$.  \exam
\subsection{The Restricted Generalized $\P$-Game}
\label{sec:restricted}
In the generalized $\cP$-game, we assumed that the bookie has
knowledge of the value of $X$ when determining the loss function $b$.
In practice, this assumption is often unrealistic.  For example, in
both the scientist (Example~\ref{ex:science}) and the weather
forecaster scenarios (Example~\ref{ex:weather}), it seems unlikely
that $b$ is in any way related to $x$.  For example, it is unlikely
that the loss function of the TV viewer is influenced by the
observations $x$ on which the weather example bases her predictions.
This motivates the {\em restricted generalized $\P$-game}, which is
just like the generalized $\P$-game, except that in step 3 of the
protocol, the value of $x$ is not observed by bookie.

Theorem~\ref{thm:semiadv} below says that, if $\Lg$ is complete, then
an agent playing against a bookie in the restricted generalized
$\P$-game should always base her decisions on a generalization of
conditioning, which we call {\em $\C$-conditioning} -- if she does
not, then the bookie can offer her bets which she is willing to
accept, but which will make her loose money on average. To formally
state the theorem, we first need to define the bookie's and agent's
strategies in the restricted generalized $\cP$-game, the notion of
$\C$-conditioning, and the requirements on the set $\cB$ that are
needed for the theorem to hold.
\paragraph{Bookie's and agent's strategies:}
The agent's strategy is a generalized decision rule $\dg \in
\dprules$, based on some probability update rule, as before. The
bookie's strategy is now a pair $(\Pr,\beta)$ with $\Pr \in \cP$ and
$\beta \in \Delta(\cB)$.  Given a generalized decision rule $\dg$, a
generalized loss function $\Lg$ and some $\beta \in \Delta(\cB)$, let
$\Lg_{\dg,\beta}$ be the random variable on $\cX \times \cY$ such that
$\Lg_{\dg,\beta}(x,y) = \sum_{b \in \cB} \beta(b) \sum_{a \in \cA}
\dg(x,b)(a) \Lg(y,a,b)$. If the bookie's move is not randomized, then
$\beta(b) = 1$ for some $b \in \cB$, and, as before, we  write
$\Lg_{\dg,b}(x,y)$.
\paragraph{$\C$-conditioning:}
 Let $\C = \{\X_1, \ldots, \X_k\}$ be
a partition of $\X$; that is, $\X_i \ne \emptyset$ for $i = 1, \ldots,
k$; $\X_1 \union \ldots \X_k = \X$; and $\X_i \inter \X_j = \emptyset$
for $i \ne j$.  If $x \in \X$, let $\C(x)$ be the cell containing $x$;
i.e., the unique element $\X_i \in \C$ such that $x \in \X_i$.  The
\emph{$\C$-conditioning} update rule is the function $\Pi$ defined by
taking $\Pi(\P,x) = \P \mid X \in \C(x)$ 
Notice that standard conditioning is  $\C$-conditioning, where we
take $\C$ to consist of all singletons.  Ignoring information is also
$\C$-conditioning, where $\C = \{\X\}$.  We say that 
A generalized
decision rule $\dg_0$ is {\em based on $\C$-conditioning\/} if it
amounts to first updating the set $\P$ to $\P \mid X \in \C(x)$ using
$\C$-conditioning, and then taking the minimax-optimal (randomized)
action relative to $\P \mid C(x)$. Formally, $\dg_0$ is based on
$\C$-conditioning if for all $x \in\cX$, all $b \in \cB$,
$$
\max_{\Pr \in \P \mid X \in C(x)} E_{\Pr}[\Lg_{\dg_0,b}]
= \min_{\dg \in \dprules} \max_{\Pr \in \P \mid X \in C(x)}
E_{\Pr}[\Lg_{\dg,b}]
$$
\edfn

We now have all ingredients to state the main theorem of this section,
which says that, if $\Lg$ is complete (Definition~\ref{def:complete}), i.e. if the bookie has the
freedom to set the prices, then the agent's minimax-optimal strategy
is to use $\C$-conditioning for some $\C$.
\begin{theorem}
\label{thm:semiadv}
Fix $\cX, \cY, \cA, \cB, \cP \subseteq \Delta(\cX \times \cY)$ and
some complete $\Lg$. Then there exists a partition $\C$ of $\cX$ such that
a generalized decision rule $(\dg)^*$ is based on
$\C$-conditioning if and only if it is a
 minimax-optimal strategy in the restricted generalized $\cP$-game, i.e.
$$
\min_{\dg \in \dprules} 
\max_{\Pr \in \cP} \max_{\beta \in \Delta(\cB)} 
E_{\Pr}[\Lg_{\dg,\beta}] = 
\max_{\Pr \in \cP} \max_{\beta \in \Delta(\cB)} 
E_{\Pr}[\Lg_{(\dg)^*,\beta}].
$$
If $\cP$ is convex and closed and $\{L(\cdot, \cdot, b) \mid b \in
\cB \}$ is convex, then the restricted generalized $\cP$-game has a
Nash equilibrium $((\Pr^*,\beta^*),(\dg)^*)$,  i.e.
\begin{equation}
\begin{array}{ccl}
E_{\Pr^*}[\Lg_{(\dg)^*,\beta^*}] & = &  
\min_{\dg \in \dprules} E_{\Pr^*}[\Lg_{\dg,\beta^*}] \\ & = &
\max_{\Pr \in \cP} \max_{\beta \in \Delta(\cB)} 
\min_{\dg \in \dprules} E_{\Pr}[\Lg_{\dg,\beta}] \\ & = &
\min_{\dg \in \dprules} 
\max_{\Pr \in \cP} \max_{\beta \in \Delta(\cB)} 
E_{\Pr}[\Lg_{\dg,\beta}] \\ & = &
\max_{\Pr \in \cP} \max_{\beta \in \Delta(\cB)} 
E_{\Pr}[\Lg_{(\dg)^*,\beta}].
\end{array}\end{equation}
\end{theorem}
\xam{\bf [Example~\ref{xam:inconsistencyb}, Cont.]}  Let $\B'$ and
$\Lg$ be the complete set of loss functions for the scenario of
Example~\ref{xam:inconsistencyb}. It turns out that in the restricted
generalized $\cP$-game, the agent's optimal decision rule $(\dg)^* \in
\dprules$ is based on ignoring the information, i.e. on
$\C$-conditioning with $\C = \{ \cX \}$.  Let us roughly sketch why
this is the case.  Note first that, while the choice of bookie's loss
function $b^*$ may depend on the agent's decision rule $(\dg)^*$, the
same $b^*$ must be chosen, irrespective of whether $X=0$ or $X=1$. By
completeness, there must be a decision rule $\delta_0$ such that
$\max_{\Pr \in \cP} E_{\Pr^*}[L_{\delta_0,b^*}] \leq 1/3$. Symmetry
considerations show that there is even a $\delta_0$ with $\delta_0(0)
= \delta_0(1)$ such that $\max_{\Pr \in \cP}
E_{\Pr^*}[L_{\delta_0,b^*}] \leq 1/3$. This implies that the agent can
ignore the value of $x$ when the bookie chooses loss function $b^*$.
Since we made no assumptions about $b^*$, it follows that by ignoring
$x$, no matter what $b^*$ is observed, the agent's expected loss is no
greater than $1/3$. By completeness of $\Lg$, the minimax value of the
game is no smaller than $1/3$.  This shows that ignoring $x$ is a
minimax-optimal strategy.  \exam

In Section~\ref{sec:game1} we showed
that agent's minimax strategy in the ordinary $\cP$-game is given by
an a priori minimax decision rule.  Theorem~\ref{thm:semiadv} shows
that in the restricted generalized $\cP$-game, the minimax strategy is
given by $\C$-conditioning for some $\cP$. 
}

\section{$\C$-conditioning and Calibration}\label{sec:calibration}

Conditioning is the most common way of updating uncertainty.  In this
section, we examine updating by conditioning.  The following definition
makes precise the idea that a decision rule is based on conditioning.

\dfn
A\label{def:ccond} \emph{probability update rule} is a function $\Pi:
  2^{\Delta(\cX 
  \times \cY)} \times \cX \rightarrow 2^{\Delta(\cX
  \times \cY)}$ mapping a set $\cP$ of distributions  and an
observation $x$ to a set $\Pi(\cP,x)$ of distributions;
intuitively, $\Pi(\cP,x)$ is 
the result of updating $\P$ with the observation $x$. 
\edfn
\dfn
Let $\C = \{\X_1, \ldots, \X_k\}$ be
a partition of $\X$; that is, $\X_i \ne \emptyset$ for $i = 1, \ldots,
k$; $\X_1 \union \ldots \X_k = \X$; and $\X_i \inter \X_j = \emptyset$
for $i \ne j$.  If $x \in \X$, let $\C(x)$ be the cell containing $x$;
i.e., the unique element $\X_i \in \C$ such that $x \in \X_i$.  The
\emph{$\C$-conditioning} 
probability
update rule is the function $\Pi$ defined by
taking $\Pi(\P,x) = \P \mid X \in \C(x)$. 
A decision rule $\delta$ is {\em based on
$\C$-conditioning\/} if it 
amounts to first updating the set $\P$ to $\P \mid X \in \C(x)$, 
and then taking the minimax-optimal 
distribution over actions
relative to $\P \mid X \in \C(x)$. 
Formally, $\delta$ is based on
$\C$-conditioning if, for all $x \in\cX$ with $\Pr(X=x) > 0$ for some
$\Pr \in \cP$, 
$$
\max_{\Pr \in \P \mid X \in \C(x)} E_{\Pr}[L_{\delta}]
= \min_{\delta \in \drules} \max_{\Pr \in \P \mid X \in \C(x)}
E_{\Pr}[L_{\delta}].
$$
\edfn

All examples of a priori minimax decision rules that we have seen so
far are based on
$\C$-conditioning: 
Standard conditioning is based on $\C$-conditioning, where we
take $\C$ to consist of all singletons;  ignoring information is also
based on $\C$-conditioning, where $\C = \{\X\}$.  
This
suggests that, perhaps, the a priori minimax decision rule 
must also be based on $\C$-conditioning. 
The following well-known example shows that this
conjecture is false.  \xam{\bf [The Monty Hall Problem]}
\label{xam:montyhall}
\cite{Mosteller,vScomb}: Suppose that you're on a game show and given
a choice of three doors.  Behind one is a car; behind the others are
goats.  You pick door 1.  Before opening door 1, Monty Hall, the host
(who knows what is behind each door) opens one of the other two doors,
say, door 3, which has a goat.  He then asks you if you still want to
take what's behind door 1, or to take what's behind door 2 instead.
Should you switch?  You may assume that initially, the car was equally
likely to be behind each of the doors.

We formalize this well-known problem as a $\cP$-game, as
follows: $\cY = \{1,2,3\}$ represents the door 
which the car is behind.  $\cX = \{G_2, G_3\}$, where, for $j
\in \{2,3\}$, $G_j$ corresponds to the quizmaster showing
that there is a goat behind door $j$. $\cA = \{1,2,3\}$, where action
$a \in \cA$ corresponds to the door you finally choose, after Monty has opened
door 2 or 3. The loss function is once again the classification loss,
$L(i,j) = 1$ if $i \neq j$, that is, if you choose a door with a goat
behind it, and $L(i,j) = 0$ if $i=j$, that is, if you choose a door with a
car.  $\cP$ is the set of all distributions $\Pr$ on $\cX \times \cY$
satisfying
$$
\begin{array}{ccc}
& \mbox{$\Pr$}_{\cY}(Y= 1) = \mbox{$\Pr$}_{\cY}(Y=2) = \mbox{$\Pr$}_{\cY}(Y=3) =
\frac{1}{3} &  \\ 
& \Pr(Y  = 2 \mid X = G_2) = 0
& \\ & 
\Pr(Y  = 3 \mid X = G_3) = 0.
&
\end{array}
$$
It is well known, and easy to show, that the minimax-optimal
strategy is always to switch doors, no matter whether Monty opens door
2 or door 3.  Since the game is an instance of the $\cP$-game, this
means that the decision rule $\delta^*$ given by
$$\delta^*(G_2) = 3 \ ;\ 
\delta^*(G_3) = 2
$$
is an a priori minimax rule. It is clear that $\delta^*$ is {\em
  not\/} based on $\C$-conditioning: there exist only two partitions
of $\cX$. The corresponding two 
update rules based on $\C$-conditioning
amount to, respectively, (a) ignoring $X$ and choosing
each door with probability 1/3, or (b) conditioning on $X$ in the
standard way and thus choosing each of the two remaining doors with
probability 1/2. Neither strategy (a) nor (b) is minimax optimal.
Thus, the a priori minimax decision rule in the $\cP$-game is not
always based on $\C$-conditioning.
\exam

While the example shows that $\C$-conditioning is not always optimal
in the minimax sense, it can be justified by other means; 
as we now show, $\C$-conditioning is closely
related to {\em calibration}. 
Indeed, a probability update rule
is calibrated if and only if 
for each $\cP$,
it amounts to ${\cal C}$-conditioning for
some partition ${\cal C}$ of ${\cal X}$.
\commentout{
In many practical situations, the analysis of Section~\ref{sec:game1}
is overly optimistic, while the analysis of
Section~\ref{sec:adversarial} is overly pessimistic. The first
analysis is overly optimistic since in practice, the agent may not
know the details of the decision problem (e.g. the loss function) in
advance.  The second analysis may be overly pessimistic because in
practice there usually is no `adversarial bookie' involved who tries
to make agent's life as hard as possible. We now study reasonable
probability update rules for such intermediate, practically relevant
cases. 
A reasonable minimum requirement for such update rules is that
they be {\em calibrated}, a notion which we now explain.
}
Calibration is usually defined in terms of empirical data. To explain
what it means, consider
a weather forecaster, who predicts the probability of rain every day.
How should we interpret the probabilities that she announces?
The usual interpretation---which
coincides with most people's intuitive understanding---%
is that, in the long run, on those days at which the weather
forecaster predict probability $p$, it will rain approximately $100 p
\%$ of the time \cite{Dawid82}. Thus, for example, among all days for
which she predicted $0.1$, the fraction of days with rain was close to
$0.1$ (given the weather forecaster's precision, we should require it
to be between, say, $0.05$ and $0.15$). A weather forecaster with this
property is said to be \emph{calibrated}. If a weather forecaster is
calibrated, and you make bets which, based on her probabilistic
predictions, seem favorable, then in the long run you cannot lose
money. If a weather forecaster is not calibrated, there exist bets
which seem favorable but which result in a loss.  
Note that calibration is a {\em minimal\/} requirement: a
weather forecaster who predicts $0.3$ for every single day of the year
may be calibrated
if it indeed rains on 30\% of the days,
but still not very informative. Thus, given two
calibrated forecasters, we prefer the one that makes  ``sharper''
predictions, in a sense to be defined below.

In our case, we do not test probabilistic predictions with respect to
empirical relative frequencies, but with respect to other sets of
``potentially underlying'' probability measures. We are not the first to
do this; see, for example, \cite{VovkGS05}. The definition of
calibration extends naturally to this situation. To see how, we first
define calibration with respect to a single underlying probability
measure.  Let $\cP = \{\Pr \}$ for a single distribution $\Pr$ and 
let $\Pi$ be a probability update rule 
(Definition~\ref{def:ccond})
such that $\Pi(\{ \Pr \},x)$ contains just a single distribution for each $x \in \cX$ (for example, $\Pi$ could be ordinary conditioning). We define 
\begin{equation}
\label{eq:range}
{\bf R} = \{ \cR : 
\cR =  ( \; \Pi(\cP,x) \;)_{\cY} \mbox{\ for some \ } x \in \cX \}.
\end{equation}
${\bf R}$ is just the range of $\Pi$, restricted to distributions of
$Y$, the random variable that we are interested in predicting; its elements are the
distributions on $Y$ that $\Pr$ is mapped to, upon observing different 
values of
$x$.
Note that ${\bf R}$ is defined relative to a 
probability update rule $\Pi$ and a set $\cP$ of distributions.
By our assumptions on $\cP$ and $\Pi$, $\cR = \{ \{R_1 \}, \{R_2 \}, \ldots
\}$ is a set of singleton sets, each containing one distribution on
$\cY$. For $\{ R \} \in {\bf R}$, let $\cX_{R}$ 
be the set of $x \in \cX$ that map $\Pr$ to $R$, i.e.
$$
\cX_{R} = \{ x \in \cX \; : \; (\; \Pi(\{ \Pr \},x) \;)_{\cY} = \{ R \}\}.
$$
Note that the sets $\{ \cX_{R} \}$ partition $\cX$. 
$\Pi$ is calibrated relative to $\cP$ if for all $R$ with $\{ R \} \in
{\bf R}$, 
$$
(\Pr \mid X \in \cX_{R} )_{\cY} = R.
$$
Thus, conditioned on the event that the agent predicts $Y$
using distribution $R$, the distribution of $Y$ must indeed be equal
to $R$.

It is straightforward to generalize this notion to sets $\cP$  of
probability distributions that are not singletons, and update rules
$\Pi$ that map to sets of 
probabilities. Definition (\ref{eq:range}) remains unchanged.  For
$\cR \in {\bf R}$, we now take $\cX_{\cR}$ to be the set of $x \in \cX$ that
map $\cP$ to $\cR$, that is,
\begin{equation}
\label{eq:image}
\cX_{\cR} = \{ x \in \cX \; : \; (\; \Pi(\cP,x) \;)_{\cY} = \cR \}.
\end{equation}
Once again, the sets $\{ \cX_{\cR} \}$ partition $\cX$. 
\dfn
\label{def:calibration}
$\Pi$ is \emph{calibrated relative to $\cP$} 
if for all 
$\Pr \in \cP$ and $\cR \in {\bf R}$,
$$
\mbox{$\Pr$}_{Y}(\cdot \mid X \in \cX_{\cR} ) \in \cR.
$$
{\em $\Pi$ is calibrated\/} if it is calibrated relative to all
sets of distributions $\cP \subseteq \Delta(\cX \times \cY)$.
\edfn

\pro\label{pro:calibration} For all partitions $\C$ of $\cX$ and all 
$\cP$, $\C$-conditioning is calibrated relative to $\cP$.
\epro

Calibration as defined here is a very weak notion. For example, the
update rule $\Pi(\cP,x) = \Delta(\cX \times \cY)$ that maps each
combination of $x$ and $\cP$ to the set of all distributions on $\cX
\times \cY$ is calibrated under our definition. This update rule
loses whatever information may have been contained in $\cP$, and is
therefore not very useful. 
Intuitively, the fewer distributions that there are in $\P$, the more
information  $\P$ contains.  Thus, we restrict ourselves to sets $\P$
that are as small as possible, while still being calibrated.
\commentout{
In order to state the main result of this section, we need two more
definitions:
\dfn
  A {\em widening\/} of an update rule $\Pi$ is an update rule $\Pi'$
  such that for all $x \in \cX$, for all sets of distributions $\cP$,
$$
\Pi(\cP,x) \subseteq \Pi'(\cP,x).
$$
If $\Pi'$ is a widening of $\Pi$, then we call $\Pi$ a {\em
  narrowing\/} of $\Pi'$.
\edfn
}

\begin{definition}
Update rule $\Pi'$ is {\em wider than update rule
  $\Pi$ relative to $\cP$} if, for all $x \in \cX$,
$$
\Pi(\cP,x) \subseteq \Pi'(\cP,x).
$$
$\Pi'$ is \emph{strictly wider} relative to $\cP$ if the inclusion is strict
for some some $x$.  
$\Pi$ is 
{\em   (strictly)  narrower\/} than $\Pi'$, relative to $\cP$  
if $\Pi'$ is (strictly) wider than $\Pi$ relative to $\cP$. 
$\Pi$ is \emph{sharply calibrated} relative to $\cP$ if $\Pi$ is
calibrated
relative to $\cP$
and there is no update rule $\Pi'$ that is calibrated 
and strictly narrower than $\Pi$ relative to $\cP$.
$\Pi$ is \emph{sharply calibrated} if $\Pi$ is sharply calibrated
relative to all $\cP \subseteq \Delta(\cX \times \cY)$.
\edfn

We now want to prove that every sharply calibrated update rule must 
involve conditioning.  To make this precise, we need the following
definition.

\dfn 
$\Pi$ is a {\em generalized conditioning update rule\/} if,
for all $\cP \subseteq \Delta(\cX \times \cY)$, there exists a partition
$\C$ (that may 
depend on $\cP$) such that for all $x \in \cX$, 
$\Pi(\cP,x) = \cP \mid C(x)$.
\edfn 
Note that in a generalized conditioning rule, we
condition on a partition of $\cX$, but the partition may depend on the
set $\P$. For example, for some $\P$, the rule may ignore the value of
$x$, whereas for other $\P$, it may amount to ordinary conditioning.
It easily follows from Proposition~\ref{pro:conditionmm} that every
generalized conditioning rule is calibrated.  The next result shows that
every \emph{sharply} calibrated update rule must be a generalized
conditioning rule.

\begin{theorem}
\label{thm:calibration}
There exists an update rule that is sharply calibrated.  Moreover,
every sharply calibrated update rule is 
a generalized conditioning update rule. 
\end{theorem}

\commentout{ \prf The if-direction is trivial. For the only-if
  direction, suppose that $\Pi$ is calibrated. Define $\phi(x) =
  \cX_{\cR}$ for the $\cR$ satisfying $x \in \cR$. (Formally, $\phi$
  is a random variable so it must map to real numbers rather than
  sets, but modification of the proof to take this into account is
  trivial). Suppose that $\Pi$ is not a widening of
  $\phi$-conditioning for this $\phi$. Then, for at least one $\cR \in
  {\bf R}$, $\cP^{(Y)}( \cdot \mid x \in \cX_{\cR})$ contains a
  distribution that is not in $\cR$. Then, by the definition of
  calibration, $\Pi$ is not calibrated, in contradiction with our
  assumption. It follows that $\Pi$ is a widening of
  $\phi$-conditioning for the $\phi$ that we defined.  \eprf 
}
Theorem~\ref{thm:calibration} says that an agent who wants to be
sharply calibrated should update her probabilities using
conditioning (although what she conditions on may depend on the set of
probabilities that she considers possible).
Given the game-theoretic interpretation of Section~\ref{sec:game1}, we
might wonder if there is a variant of the games considered earlier for which
the equilibrium 
involves generalized conditioning.
As we show in the full
paper, there is (although the game is perhaps not as natural as the ones
considered in Section~\ref{sec:game1}).  Roughly speaking, we consider a
three-player game, with a bookie and two agents.  The bookie again
chooses a probability distribution from a set $\P$;
the bookie also chooses the loss function from some set.
The first agent observes $\P$ and $x$ and updates $\P$ to $\P_x$.  The
second agent learns $\P_x$ and $b$ (but not $\P$ and $x$) and makes the
minimax-optimal decision.  As we show, in Nash equilibrium, the first
agent's updated set of probabilities, $\P_x$, must be the result of
$\C$-conditioning, where, as in Theorem~\ref{thm:calibration}, $\C$ may
depend on $\P$. 
\commentout{
In practice, she will not be interested in strict 
widenings of $\phi$-conditionings, since these sometimes perform worse, 
but never perform better than $\phi$-conditioning itself:
\begin{proposition}
  Fix some $\phi: \cX \rightarrow \reals$. Let $\cS$ be the set of
  $\Pi$-based decision rules, where $\Pi$ is a widening of
  $\phi$-conditioning. Let $\dg_\phi \in \cS$ be the decision rule
  based on $\phi$-conditioning (without widening). Then $\dg_\phi$
  dominates $\cS$, in the sense that for all $\dg \in \cS$, {\em all\/} $\Pr \in \cP$,
{\em all\/} sets of loss functions $\Lg(\cdot,\cdot,b), b \in \cB$ 
and {\em  all\/} generalized decision rules $\dg \in \cS$,
$$
E_{\Pr}[\Lg(Y,\dg_{\phi}(X,b),b)] \leq 
E_{\Pr}[\Lg(Y,\dg(X,b),b)]. 
$$
\end{proposition}
}
\section{Discussion}
\label{sec:discussion}
We have examined how to update uncertainty represented by a set of
probability distributions,
where we motivate updating rules in terms of the minimax criterion.
Our key innovation 
has been to show
how different approaches can be understood in terms of a game between a
bookie and an agent, where the bookie picks a distribution from the
set and the agent chooses an action after making an observation.
Different approaches to updating arise depending on whether the bookie's
choice is made before or after the observation.  
We believe that this game-theoretic approach should prove useful more
generally in understanding different approaches to updating.  We hope to
explore this further in future work.

We end this paper by giving an overview of the senses in which
conditioning is optimal and the senses in which it is not, when
uncertainty is represented by a set of distributions. We have
established that  conditioning the full set $\cP$ on $X=x$ is minimax optimal
in the $\cP$-$x$-game, but not in the $\cP$-game. The minimax-optimal
decision rule in the $\cP$-game is often an instance of 
$\C$-conditioning, a generalization of conditioning. The Monty
Hall problem showed, however, that this is not always the case. 
On the other hand, if instead of the minimax criterion, we insist that
update rules are 
calibrated, then $\C$-conditioning is always the right thing to do
after all. 

There are two more senses in which conditioning is the
right thing to do.
First, 
Walley \citeyear{Walley91} shows that, in a sense,
conditioning is the only updating rule that is \emph{coherent},
according to his notion of coherence.  He justifies coherence decision
theoretically, but not by using the minimax criterion.  
Note that the minimax criterion puts a total order on decision
rules.  That is, we can say that $\delta$ is at least as good as $\delta'$ if 
$${\max}_{\Pr \in \P} E_{\Pr}[L_\delta] \le {\max}_{\Pr \in \P}
E_{\Pr}[L_{\delta'}].$$
By way of contrast, Walley \citeyear{Walley91} puts a partial order on
decision rules by taking $\delta$ to be at least as good as $\delta'$ if
$${\max}_{\Pr \in \P} E_{\Pr}[L_\delta - L_{\delta'}] \le 0.$$  
Since both ${\max}_{\Pr \in \P} E_{\Pr}[L_\delta - L_{\delta'}]$ and 
${\max}_{\Pr \in \P} E_{\Pr}[L_{\delta'} - L_{\delta}]$ may be
positive, this is indeed a partial order.  If we use this ordering to
determine the optimal decision rule then, as Walley shows, conditioning
is the only right thing to do.

Second, in this paper, we interpreted ``conditioning''
as conditioning the full given set of distributions $\cP$. Then
conditioning is not always an a priori minimax optimal strategy
on the observation $X=x$.
Alternatively, we could first somehow select a {\em single\/} $\Pr \in 
\cP$, condition $\Pr$ on the observed $X=x$, and then take the optimal
action relative to $\Pr \mid X=x$. 
It follows from Theorem~\ref{thm:dawida} that the minimax-optimal
decision rule $\delta^*$ in a $\cP$-game can be understood this way.
It defines the optimal response to the distribution $\Pr^* 
\in \Delta(\cX \times \cY)$
defined in Theorem~\ref{thm:dawida}(b)(ii). If
$\cP$ is convex, then $\Pr^* \in \cP$. 
In this sense, the minimax-optimal
decision rule can always be viewed as an instance of ``conditioning,''
but on a
single special $\Pr^*$ that  depends on the loss function $L$ rather
than on the full set $\cP$. 
It is worth noting that Grove and Halpern \citeyear{GroveHalpern98}
give an axiomatic characterization of conditioning sets of
probabilities, based on axioms given by van Fraassen \citeyear{vF1,vF3}
that characterizing conditioning in the case that uncertainty is
characterized by a single probability measure.  
As Grove and Halpern point out, their axioms are not as compelling as
those of van Fraassen.  It would be interesting to know whether an 
axiomatization that is similar in spirit can be used to characterize the
update notions that we have considered here.

\commentout{
\section*{Acknowledgements}
We thank Teddy Seidenfeld and Bas van Fraassen for helpful discussions
on the topic of the paper.  Peter Gr\"unwald was supported by the IST Programme of the European
Community, under the PASCAL Network of Excellence,
IST-2002-506778. Joseph Halpern was supported in part by
NSF under grants CTC-0208535 and ITR-0325453, by ONR under grants
N00014-00-1-03-41 and N00014-01-10-511, by the DoD Multidisciplinary
University Research Initiative (MURI) program administered by the ONR
under grant N00014-01-1-0795, and by AFOSR under 
grants F49620-02-1-0101 and FA9550-05-1-0055.
This publication reflects only the authors' views.
}

\bibliographystyle{chicago}
\bibliography{z,joe,bghk,refs}

\appendix

\section{Proofs}
\label{sec:proofs}
\commentout{
JOE, A GENERAL ISSUE THAT CAME UP IS THE FOLLOWING: for general $\cP$
that are {\em not\/} closed, the a priori/a posteriori minimax
decision rules can be undefined because $\max_{\Pr \in \cP} E_{\Pr} 
[L_{\delta}]$ may not be achieved. We should either say something
about this, or always assume that $\cP$ is closed (we don't do that
know), or, perhaps better, replace the $\max$ in the definition
by a $\sup$, i.e. the definition becomes
$\sup_{\Pr \in \cP} E_{\Pr}[L_{\delta^*}] = 
\min_{\delta \in \drules} \sup_{\Pr \in \cP} E_{\Pr}[L_{\delta}]$.
With that definition the minimax decision rule is {\em always\/} defined:
note that the $\min$ is not problematic, because $\drules$ is always a
closed set.
END OF GENERAL REMARK
}
To prove Theorems~\ref{thm:dawida} and Theorem~\ref{thm:posteriorb}, we
need two preliminary observations.  The first is a characterization of
Nash equilibria.
In the $\cP$-game, a Nash equilibrium or saddle point amounts to  a pair
$(\pi^*,\delta^*)$ where $\pi^*$ is a distribution on $\cP$ and
$\delta^*$ is a randomized decision rule such that
\begin{equation}
\label{eq:spbasic}
\begin{array}{cl}
& E_{\pi^*} E_{\Pr} [L_{\delta^*}] \\
= & 
{\min}_{\delta \in \drules} 
E_{\pi^*} E_{\Pr} [L_{\delta}]\\  
=  &{\max}_{\Pr \in \cP} E_{\Pr} [L_{\delta^*}],
\end{array}
\end{equation}
where $E_{\pi^*} E_{\Pr}$ is an abbreviation of $\sum_{\Pr \in \cP,
\pi^*(\Pr) > 0} \pi^*(\Pr) \Pr$. In the $\cP$-$x$-game, a 
Nash equilibrium
is a pair $(\pi^*,\delta^*)$ where $\pi^*$ is a distribution on $\cP
\mid X=x$ and
$\delta^*$ is a randomized decision rule, such that (\ref{eq:spbasic})
holds with $\cP$ replaced by $\cP \mid X=x$.

The second observation we need is the following special case of Theorem
3.2 in  \cite{GrunwaldD04}, itself an
extension of Von Neumann's original minimax theorem. 
\begin{theorem}
\label{thm:dawid52}
If $\cY'$ is a finite set, $\cP'$ is a closed and convex subset
of $\Delta(\cY')$, $\cA'$
a closed and convex subset of $\reals^{k}$ for some $k \in \IN$, and
$L': \cY' \times 
\cA' \rightarrow \reals$ is a bounded function such that, for each $y
\in \cY'$, $L(y,a)$ is a continuous function of $a$, then there exists
some $\Pr^* \in \cP'$ and some $\rho^* \in \cA'$ such that,
\begin{equation}
\label{eq:spa}
\begin{array}{cl}
& E_{\Pr^*} [L'(Y',\rho^*)] \\
= & 
{\min}_{\rho \in \cA'} 
E_{\Pr^*} [L'(Y',\rho)]\\  
=  &{\max}_{\Pr \in \cP'} E_{\Pr} [L'(Y',\rho^*)].
\end{array}
\end{equation}
\end{theorem}

With these observations, we are ready to prove Theorem~\ref{thm:dawida}:

\othm{thm:dawida}
Fix $\cX$, $\cY$, $\cA$, $L$, and $\cP \subseteq \Delta(\cX \times \cY)$.
\begin{itemize}
\item[(a)] The $\P$-game has a Nash equilibrium  
$(\pi^*,\delta^*)$, where $\pi^*$ is a distribution over $\P$ with
finite support.
\item[(b)] If $(\pi^*,\delta^*)$ is a Nash equilibrium of the
$\P$-game
such that $\pi^*$ has finite support,
then 
\begin{itemize}
\item[(i)] 
for every distribution $\Pr' \in \P$ in the support of $\pi^*$, 
 we have
$$
E_{\Pr'} [L_{\delta^*}] = 
{\max}_{\Pr \in \cP} E_{\Pr} [L_{\delta^*}];
$$ 
\item[(ii)] if $\Pr^* = \sum_{\Pr \in \cP, \pi^*(\Pr) > 0} \pi^*(\Pr)
\Pr$ (i.e., $\Pr^*$ is the convex combination of the distributions in
the support of $\pi^*$, weighted by their probability according to
$\pi^*$), then 
$$
\begin{array}{cl}
&E_{\Pr^*} [L_{\delta^*}]\\
= & 
{\min}_{\delta \in \drules} 
E_{\Pr^*} [L_{\delta}]\\  
= &\max_{\Pr \in \cP} {\min}_{\delta \in \drules} 
E_{\Pr} [L_{\delta}] \\
= &  
{\min}_{\delta \in \drules} 
{\max}_{\Pr \in \cP} E_{\Pr} [L_{\delta}] \\
= &{\max}_{\Pr \in \cP} E_{\Pr} [L_{\delta^*}].
\end{array}
$$ 
\end{itemize}
\end{itemize}
\eothm

\prf
To prove part (a), we introduce a new loss function $L'$ that is
essentially equivalent to $L$, but is designed so that
Theorem~\ref{thm:dawid52} can be applied. Let $\cY' = \cX \times
\cY$, let $\cA' = \drules$, and
define the function $L': \cY' \times \cA' \rightarrow \reals$ as
$$
L'((x,y),\delta) := L_{\delta}(x,y) = 
\sum_{a \in \cA} \delta(x)(a)L(y,a).
$$
Obviously $L'$ is equivalent to $L$ in the sense
that for all $\Pr \in \Delta(\cX \times \cY)$, for all $\delta \in \drules$,
$$
E_{\Pr}[L_{\delta}] = E_{\Pr}[ L'((X,Y),\delta)].
$$
If we view $\cA' = \drules$  as a convex subset of
$\reals^{| \cX| \cdot (| \cA|-1)}$, then $L'((x,y),a)$ becomes a
continuous function of $a \in \cA'$. Let  $\cP'$
be the convex closure of $\cP$.  
Since $\cX \times \cY$ is finite, $\cP'$ consists of all distributions
$\Pr^*$ on $(\cX,\cY)$ of the form $c_1 \Pr_1 + \cdots + c_k \Pr_k$ for
$k = |\cX \times \cY|$, 
where $\Pr_1, \ldots, \Pr_k \in \cP$ and $c_1,
\ldots, c_k$ are nonnegative real
coefficients such that $c_1 + \cdots + c_k = 1$.  
Applying 
Theorem~\ref{thm:dawid52} to $L'$ and $\cP'$, it follows that
(\ref{eq:spa}) holds for  
some $\Pr^* \in \cP'$ and some $\delta^* \in \cA' = \drules$ (that is,
the $\rho^*$ in (\ref{eq:spa}) is $\delta^*$).  
Thus,
there must be some distribution $\pi^*$ on 
$\cP$ with finite support such that $\Pr^* = \sum_{\Pr \in \cP,
  \pi^*(\Pr) > 0} \pi^*(\Pr) \Pr$. 
It is easy to see that the two equalities in (\ref{eq:spa}) are
literally the two equalities in (\ref{eq:spbasic}).  Thus, 
$(\pi^*,\delta^*)$ is a Nash equilibrium. This proves part (a).

To prove part (b)(i), suppose
first that $(\pi^*,\delta^*)$ is a Nash equilibrium of the $\cP$-game
such that $\pi^*$ has finite support.
Let $V = \max_{\Pr \in \cP} E_{\Pr}[L_{\delta^*}]$. By
(\ref{eq:spbasic}), we have that
\begin{equation}
\label{eq:value}
\sum_{\Pr \in \cP, \pi^*(\Pr) > 0} \pi^*(\Pr) E_{\Pr}[L_{\delta^*}] = V.
\end{equation}
Trivially, for each $\Pr' \in \cP$, we must have $E_{\Pr'}[L_{\delta^*}] \leq
\max_{\Pr \in \cP} E_{\Pr}[L_{\delta^*}]$.  If this inequality
were strict for some $\Pr' \in \cP$ in the support of $\pi^*$, then $
\sum_{\Pr \in \cP, \pi^*(\Pr) > 0} \pi^*(\Pr) E_{\Pr}[L_{\delta^*}] <
V$, contradicting (\ref{eq:value}). This proves part (b)(i).

To prove part (b)(ii), note that straightforward arguments show that
$$
\begin{array}{cl}
& \max_{\Pr \in \cP} E_{\Pr} [L_{\delta^*}] 
\\ \geq & \min_{\delta \in \drules}
\max_{\Pr \in \cP} E_{\Pr} [L_{\delta}] \\
\geq & \max_{\Pr \in \cP} 
{\min}_{\delta \in \drules} E_{\Pr} [L_{\delta}] \\
\geq & {\min}_{\delta \in \drules} 
E_{\Pr^*} [L_{\delta}].
\end{array}
$$
(The second inequality follows because, for all $\Pr' \in \cP$, 
$\min_{\delta \in \drules} \max_{\Pr \in \cP} E_{\Pr} [L_{\delta}] 
\geq  
{\min}_{\delta \in \drules} E_{\Pr'} [L_{\delta}]$.)
Since $(\pi^*,\delta^*)$ is a Nash equilibrium, part (b)(ii) is
immediate, using the equalities in (\ref{eq:spbasic}).
\eprf 

\othm{thm:posteriorb} 
Fix $\cX$, $\cY$, $\cA$, $L$, $\cP \subseteq \Delta(\cX \times   \cY)$. 
\begin{itemize}
\item[(a)] The $\P$-$x$-game has a Nash equilibrium
$(\pi^*,\delta^*(x))$, where $\pi^*$ is a distribution over $\cP \mid
X=x$ with finite support.
\item[(b)] If $(\pi^*,\delta^*(x))$ is a Nash equilibrium of the
$\P$-$x$-game
such that $\pi^*$ has finite support, 
then 
\begin{itemize}
\item[(i)]
for all $\Pr'$ in the support of $\pi^*$, we have
$$E_{\Pr'} [L_{\delta^*}] = 
{\max}_{\Pr \in \cP \mid X=x} E_{\Pr} [L_{\delta^*}];$$
\item[(ii)] if $\Pr^* = \sum_{\Pr \in \cP, \pi^*(\Pr) > 0} \pi^*(\Pr)
\Pr$, then 
$$
\begin{array}{cl}
&E_{\Pr^*} [L_{\delta^*}]\\
= & 
{\min}_{\delta \in \drules} 
E_{\Pr^*} [L_{\delta}]  \\

= &\max_{\Pr \in \cP \mid X=x} {\min}_{\delta \in \drules} 
E_{\Pr} [L_{\delta}] \\
= &  
{\min}_{\delta \in \drules} 
{\max}_{\Pr \in \cP \mid X=x} E_{\Pr} [L_{\delta}] \\
= &{\max}_{\Pr \in \cP \mid X=x} E_{\Pr} [L_{\delta^*}].
\end{array}
$$
\end{itemize}
\end{itemize}
\eothm
\prf
To prove part (a), we apply Theorem~\ref{thm:dawid52}, 
setting $L' = L$, $\cY' = \cY$,
$\cA' = \Delta(\cA)$, and $\cP'$ to the convex closure of $\cP \mid
X=x$. 
Thus,
(\ref{eq:spa}) holds for some $\rho^* \in \cA'$, which
we denote $\delta^*(x)$. 
As in the proof of Theorem~\ref{thm:dawida}, 
there must be some distribution $\pi^*$ on 
$\cP \mid X=x$ with finite support such that 
$\Pr^* = \sum_{\Pr \in \cP \mid X=x,   \pi^*(\Pr) > 0} \pi^*(\Pr) \Pr$. 
The remainder of the argument is identical to that in
Theorem~\ref{thm:dawida}.

The proof of part (b) is completely analogous to the proof of part
(b) of Theorem~\ref{thm:dawida}, and is thus omitted.  \eprf

\opro{pro:conditionmm} If $\cP = \hull{\cP}$, then there exists an a priori
minimax-optimal rule that is also a posteriori minimax optimal. If, for
all $\Pr \in \cP$ and all $x \in \cX$, $\Pr(X=x) > 0$, then every a
priori minimax-optimal rule is also a posteriori minimax optimal. \eopro

\prf 
\commentout{
Suppose that $\delta$ is an a priori minimax-optimal decision
rule.  
Let $\cP_\delta = \{ \Pr \in \Delta(\cX \times \cY) 
\;:\; E_{\Pr}[L_{\delta}]=
\max_{\Pr' \in \cP} E_{\Pr'} [ L_{\delta}]$.  Since $\cP$ is closed,
$\cP_\delta$ is nonempty.
Let $\cX' = \{x \in \cX: \max_{\Pr \in \cP_\delta} \Pr(X=x) > 0\}$.
Let $\delta^*$ be such that $\delta^*(x) = \delta(x)$ if $x \in \cX'$,
and $\delta'(x)$ is minimax optimal with respect to $\cP \mid X= x$.  We
now show 
that $\delta^*$ is both a priori minimax optimal and a posteriori
minimax optimal.  This suffices to prove the result since if $\Pr(X=x) >
0$ for all $\Pr t\in \cP$ and all $x \in \cX$, then $\cX' = \cX$, so it
follows that every a priori minimal-optimal decision rule is also a
posteriori minimal optimal.

To show that $\delta^*$ is a priori minimax optimal, note that for all
$\Pr \in \cP_\delta$, we have $E_{\Pr}[L_\delta] =
E_{\Pr}[L_{\delta^*}]$, since 
$\delta^*(x) = \delta(x)$ for all $x$ such that $\Pr(X=x) > 0$.  Thus, we
must have $\max_{\Pr' \in \cP} E_{\Pr'} [ L_{\delta^*}] \ge
\max_{\Pr' \in \cP} E_{\Pr'} [ L_{\delta}]$.  
Since
$\delta$ is minimax optimal, it easily follows that $\delta^*$ is as
well.

To see that $\delta^*$ is a posteriori minimax optimal, it suffices to
show that for all $x \in \cX'$, $\delta(x)$ is minimax optimal with
respect to $\cP \mid X = x$.  
For each $x \in \cX'$, 
choose $\Pr_x \in \cP^*$ such that $\Pr_x(X=x) > 0$.  Since $\cX$ is
finite and $\cP^*$ is convex, there is an element $\Pr^*x \in \cP^*$
(which can be obtained by taking a convex combination of the
distribution $\Pr_x$ such that, for all $x \in \cX$, 
$$
\mbox{if $\Pr^*(X=x) = 0$ then $\max_{\Pr \in \cP^*} \Pr(X=x) = 0$}.
$$
For each $x \in \cX'$, 
let $\Pr^x \in \cP$ be such that $E_{\Pr^x}[L_{\delta^*}] = 
\max_{\Pr \in \cP  \mid X=x} E_{\Pr}[L_{\delta^*}]$.  
If $x \notin \cX'$ but $\cP \mid X=x$ is nonempty, let $\Pr^x$ be an
arbitrary element of $\cP$ such that $\Pr(X = x) > 0$.
Finally, if $\cP \mid X=x$ is
empty, let $\Pr^x$ be an arbitrary element of $\Delta(\cY)$.
Define $\Pr^{\dag} \in \Delta(\cX \times \cY)$ by taking
$\Pr^{\dag}(x,y) =  \Pr^*_{\cX}(X=x) \Pr^x(Y=y)$.  Clearly $\Pr^{\dag}_{\cX} =
\Pr^*_{\cX}$ and $\Pr^{\dag} \mid X=x = \Pr^x \mid X = x \in \cP \mid X=x$
if $\Pr \mid X =x$ is nonempty.
Thus, by definition, $\Pr^{\dag} \in \hull{\cP}$.  Since, by
assumption, $\hull{\cP} = \cP$, it follows that $\Pr^{\dag} \in \cP$.
We claim that $\Pr^{\dag} \in \cP^*$.  To see this, note that, by
definition of $\Pr_x$ and $\Pr^{dag}$,
$$\begin{array}{ll}
&E_{\Pr^*}[L_{\delta^*}]\\
= & \sum_{(x,y) \in \cX \times \cY} \Pr^*(x,y) L_{\delta^*}(x,y)\\
= &\sum_{\{x \in X: \Pr^*_{\cX}(X=x) > 0\}} \Pr^*_{\cX}(X=x) \sum_{y \in \cY}
\Pr^*(Y = y \mid X=x) L_{\delta^*}(x,y)\\
\le &\sum_{\{x \in X: \Pr^*_{\cX}(X=x) > 0\}} \Pr^*_{\cX}(X=x) \sum_{y \in \cY}
\Pr_x(Y = y \mid X=x) L_{\delta^*}(x,y)\\
= &\sum_{\{x \in X: \Pr^\dag_{\cX}(X=x) > 0\}} \Pr^\dag_{\cX}(X=x) \sum_{y \in \cY}
\Pr^\dag(Y = y \mid X=x) L_{\delta^*}(x,y)\\
= &E_{\Pr^\dag}[L_{\delta^*}].
\end{array}
$$

It follows that 
\begin{equation}
\label{eq:sollyb}
\begin{array}{cl}
& \max_{\Pr \in \cP} E_{\Pr}[L_{\delta^*}] \\
= & 
\sum_{x \in \cX} \Pr^\dag_{\cX}(X=x)
\cdot \left( \max_{\Pr' \in \cP \mid X = x} E_{\Pr'}[L_{\delta^*}] 
\right)
\end{array}
\end{equation}
Since $\delta^*$ is a priori minimax optimal,
$$
\max_{\Pr \in \cP} E_{\Pr}[L_{\delta^*}] 
=  \min_{\delta \in \drules} 
\max_{\Pr \in \cP} E_{\Pr}[L_{\delta}],
$$
so that, with (\ref{eq:sollyb}), we get
$$
\begin{array}{l}
\sum_{x \in \cX} \Pr^\dag_{\cX}(X=x)
\cdot \left( \max_{\Pr' \in \cP \mid X = x} E_{\Pr'}[L_{\delta^*}] 
\right) = \\ 
\min_{\delta \in \drules} 
\sum_{x \in \cX} \Pr^\dag_{\cX}(X=x) \cdot \left(
\max_{\Pr' \in \cP \mid X = x} E_{\Pr'}[L_{\delta}] \right).
\end{array}
$$
This implies that $\delta^*(x)$ is minimax optimal in the
$\cP$-$x$-game for all $x \in \cX'$.  
}
Let $\cX^+ = \{x \in \cX: \max_{\Pr \in \cP} \Pr(X=x)> 0 \}$. 
Let $m_\delta$ be a random variable on $\cX$ defined by taking
$m_\delta(x) = 0$ if $x \notin \cX^+$, 
and $m_\delta(x) =
\max_{\Pr' \in \cP \mid X=x} E_{\Pr'}[L_\delta]$ if $x \in \cX^+$.
We first show that for every $\delta \in \drules$,
\begin{equation}
\label{eq:eurandom}
\max_{\Pr \in \cP} E_{\Pr} [ L_\delta] = 
\max_{\Pr \in \cP} \sum_{x \in \cX} {\Pr}_{\cX}(X=x) m_\delta(x).
\end{equation}
Note that 
$$\begin{array}{ll}
&E_{\Pr}[L_\delta] \\
= &\sum_{(x,y) \in \cX \times \cY} \Pr((X,Y) = (x,y)) L_\delta(x,y)\\
= &\sum_{\{x \in \cX: \Pr_\cX(x) > 0\}} \Pr_{\cX}(X=x)
\sum_{y \in \cY} \Pr(Y=x \mid X =x) L_\delta(x,y)\\
= &\sum_{\{x \in \cX: \Pr_\cX(x) > 0\}} \Pr_{\cX}(X=x) 
E_{\Pr \mid X=x}[L_{\delta}]\\
\le &\sum_{\{x \in \cX: \Pr_\cX(x) > 0\}} \Pr_{\cX}(X=x)
\max_{\Pr' \in \cP \mid X=x} E_{\Pr'}[L_{\delta}]\\
=  &\sum_{\{x \in \cX: \Pr_\cX(x) > 0\}} \Pr_{\cX}(X=x)
m_\delta(x)\\
=  &\sum_{x \in \cX} \Pr_{\cX}(X=x)
m_\delta(x).
\end{array}$$

Taking the max over all $\Pr \in \cP$, we get that 
$$\max_{\Pr \in \cP} E_{\Pr} [ L_\delta] \le  \max_{\Pr \in \cP} \sum_{x
\in \cX} {\Pr}_{\cX}(X=x) m_\delta(x).$$

It remains to
show the reverse inequality in (\ref{eq:eurandom}). 
Since
$\cP$ is closed, 
there exists $\Pr^* \in \cP$ such that 
$$\max_{\Pr \in \cP}
\sum_{x \in \cX} {\Pr}_{\cX}(X=x) m_\delta(x) = 
\sum_{x \in \cX} {\Pr}^*_{\cX}(X=x) m_\delta(x).$$ 
Moreover, since $\cP \mid  X = x$ is closed, if $x \in \cX^+$, 
there exists $\Pr^x \in \cP \mid X=x$ such that 
$m_\delta(x) = E_{\Pr^x}[L_\delta]$.  Define
 $\Pr^{\dag} \in \Delta(\cX \times \cY)$ by taking
$$
{\Pr}^{\dag}((X,Y) = (x,y)) =  
\left\{
\begin{array}{lll}
0 &\mbox{if $x \notin \cX^+$}\\
{\Pr}^*_{\cX}(X=x) \Pr^x(Y=y) &\mbox{if $x \in
\cX^+$.}
\end{array}\right.
$$  Clearly 
$\Pr^{\dag}_{\cX} = 
\Pr^*_{\cX}$ and $(\Pr^{\dag} \mid X=x) = (\Pr^x \mid X = x) \in \cP
\mid X=x$ 
if $x \in \cX^+$.  Thus, by definition, $\Pr^{\dag} \in
\hull{\cP}$.  Since, by assumption, $\hull{\cP} = \cP$, it follows that
$\Pr^{\dag} \in \cP$.   
In addition,
it easily follows that
$$\begin{array}{ll}
&\max_{\Pr \in \cP}  \sum_{x \in \cX}
\Pr_\cX(X=x) m_\delta(x) \\
= &\sum_{x \in \cX} \Pr^\dag_{\cX}(X=x) m_\delta(x) \\
= &\sum_{x \in \cX^+} \Pr^\dag_{\cX}(X=x) \sum_{y \in \cY} \Pr^\dag(Y =
y \mid X = x) L_\delta(x,y)\\
= &E_{\Pr^\dag}[L_\delta] \\
\leq &\max_{\Pr \in \cP} E_{\Pr}[L_{\delta}].
\end{array}
$$
This establishes (\ref{eq:eurandom}).

Now let $\delta^*$ be an a priori minimax decision rule. Since the
$\cP$-game has a Nash equilibrium (Theorem~\ref{thm:dawida}), such a
$\delta^*$ must exist. Let $\cX'$ be the set of all $x' \in \cX$ for
which $\delta^*$ is not minimax optimal in the $\cP$--$x'$-game, i.e.,
$x' \in \cX'$ iff $x \in \cX^+$ and $\max_{\Pr' \in
  \cP \mid X=x'} E_{\Pr'}[L_{\delta^*}] > \min_{\delta \in \drules}
\max_{\Pr' \in \cP \mid X=x'} E_{\Pr'}[L_{\delta}]$.  
Define $\delta'$ to be a decision rule that agrees with $\delta^*$ on
$\cX \setminus \cX'$
and is minimax optimal in the $\cP \mid X = x'$ game for all $x'
\in \cX'$; that is, 
$\delta'(x) = \delta(x)$ for $x \notin \cX'$ and, for $x \in \cX'$,
$$\delta(x) \in \argmin_{\delta \in \drules} \max_{\Pr' \in \cP 
  \mid X=x'} E_{\Pr'}[L_{\delta}].$$
By construction, $m_{\delta'}(x) \le m_{\delta^*}(x)$ for all $x \in
\cX$ and $m_{\delta'}(x) < m_{\delta^*}(x)$ for all $x \in \cX'$.
Thus, using (\ref{eq:eurandom}), we have
\commentout{
\begin{eqnarray}
\label{eq:eurandomb}
& \max_{\Pr \in \cP}  E_{\Pr}[L_{\delta'}]
& \max_{\Pr \in \cP_{\cX}} \biggl( \sum_{x \in \cX \setminus \cX'} \Pr(X=x) 
m_x(\delta^*) + & \nonumber \\
& \sum_{x \in \cX'} \Pr(X=x) (m_x(\delta^*) - \epsilon) \biggr) & \nonumber
\\
& \leq \max_{\Pr \in \cP} [L_{\delta^*}], & 
\end{eqnarray} 
}%
\begin{equation}
\label{eq:eurandomb}
\begin{array}{ll}
&\max_{\Pr \in \cP}  E_{\Pr}[L_{\delta'}]\\
= & \max_{\Pr \in \cP} \sum_{x \in \cX} \Pr(X=x) m_{\delta'}(x)\\
\le & \max_{\Pr \in \cP} \sum_{x \in \cX} \Pr(X=x) m_{\delta^*}(x)\\
= &\max_{\Pr \in \cP}  E_{\Pr}[L_{\delta^*}].
\end{array}
\end{equation}
Thus,  $\delta'$ is also an a priori minimax decision rule.  But, by
construction, $\delta'$ is also an a posteriori minimax decision rule,
and it follows that there exists at least one decision rule (namely,
$\delta'$) that is both a priori and a posteriori minimax optimal. 
Moreover, if $\Pr(X=x) > 0$ for all $\Pr \in \cP$ and $x \in \cX$ and
$\cX' \ne \emptyset$,
then the inequality in (\ref{eq:eurandomb}) is strict. 
It follows that  $\cX'$ is empty in this case, for otherwise $\delta^*$
would not be a priori minimax optimal,
contradicting our assumptions. But, if $\cX'$ is empty, then $\delta^*$
must also be a posteriori minimax optimal.  \eprf 

\othm{thm:ignoremm} Fix $\cX$, $\cY$, $L$, $\cA$, and $\cP \subseteq
\Delta(\cX \times \cY)$.  
If, for all $\Py \in
\cP_{\cY}$, $\cP$ contains a distribution $\Pr'$ such that $X$ and $Y$ are
independent under $\Pr'$, and $\Pr'_{\cY} = \Py$, then there is an a
priori minimax-optimal decision rule that ignores information.  
Under these conditions,
if $\delta$ is an a priori minimax-optimal decision rule
that ignores information, then $\delta$ essentially optimizes with
respect to the marginal on $Y$; that is, $\max_{\Pr \in \cP}
E_{\Pr}[L_\delta] = \max_{\Pr_{\cY} \in \cP_{\cY}} E_{\Pr_{\cY}}[L'_{\delta}]$.
\eothm 
\prf Let $\cP'$ be the subset of $\cP$ of distributions under
which $X$ and $Y$ are independent. Let $\drules'$ be the subset of
$\drules$ of rules that ignore information. Let $\delta^* \in
\drules'$ be defined as the optimal decision rule that ignores information
relative to $\cP'$, i.e. 
$$
\max_{\Pr \in \cP'}  E_{\Pr} [L_{\delta^*}] = 
\min_{\delta \in \drules'} \max_{\Pr \in \cP'} E_{\Pr}[L_{\delta}].
$$
We have
\begin{equation}
\label{eq:mmoncemore}
\begin{array}{clr}
& \max_{\Pr \in \cP}  E_{\Pr} [L_{\delta^*}] \\
\geq & \min_{\delta \in \drules} \max_{\Pr \in \cP} E_{\Pr}[L_{\delta}] \\ 
\geq & 
\min_{\delta \in \drules} \max_{\Pr \in \cP'} E_{\Pr}[L_{\delta}] \\
= & \min_{\delta \in \drules'} \max_{\Pr \in \cP'} E_{\Pr}[L_{\delta}]
&\mbox{[see below]}\\
= & \max_{\Pr \in \cP'}  E_{\Pr} [L_{\delta^*}] .
\end{array}
\end{equation}
To see that the equality between the third and fourth line in
(\ref{eq:mmoncemore}) holds, note that for $\Pr \in \cP'$, we have 
$$\begin{array}{ll}
&E_{\Pr}[L_{\delta}]\\
= &\sum_{(x,y) \in \cX \times \cY} \Pr(x,y) L_{\delta}(x,y)\\
= &\sum_{x\in \cX} \Pr(X=x) \sum_{y \in \cY} \Pr(Y = y) (\sum_{a \in A}
\delta(x)(a) L(y,a))
\end{array}
$$
The decision rule that minimizes 
this expression
is independent of $x$; it is the
distribution $\delta^*$ over actions that minimizes
$$\sum_{y \in \cY} \Pr(Y = y) (\sum_{a \in A}
\delta^*(a) L(y,a)).$$

This calculation also shows that, since $\delta^*$
ignores information, for $\Pr \in \cP'$, we have that
\begin{equation}
\label{eq:ignorex}
\max_{\Pr \in \cP}  E_{\Pr} [L_{\delta^*}] =
\max_{\Pr_{\cY} \in \cP_{\cY}}  E_{\Pr_{\cY}} [L'_{\delta^*}] = 
\max_{\Pr \in \cP'}  E_{\Pr} [L_{\delta^*}].
\end{equation}
This implies that the first and last line of (\ref{eq:mmoncemore}) are
equal to each other, and therefore also equal to the second line of
(\ref{eq:mmoncemore}). It follows that 
$\delta^*$ is a priori minimax optimal. Since every a priori minimax
optimal rule that ignores information must satisfy
(\ref{eq:ignorex}), the second result follows.
\eprf

\opro{pro:calibration} For all partitions $\C$ of $\cX$ and all 
$\cP$, $\C$-conditioning is calibrated relative to $\cP$.
\eopro

\prf
Let $\C = \{ \cX_1, \ldots, \cX_k\}$ be a partition consisting of $k
\geq 1$ elements. 
Let
$\cR_j = (\cP \mid X \in \cX_j)_{\cY}$ and let $\cX_{\cR_j} = \cX_j$, for $j =
1, \ldots, k$; let
${\bf R} = \{ \cR_1, \ldots , \cR_k \}$. Plugging this into
Definition~\ref{def:calibration}, we find that for $\C$-conditioning
to be calibrated, we must have 
that 
$\Pr(\cdot \mid X \in \cX_j) \in \cP \mid X \in \cX_j$
for all $\cP \subseteq \Delta(\cX \times \cY)$, all $\Pr \in\cP$, and all 
$ j \in \{1, \ldots, k\}$. 
But this is true by definition of $\cP \mid X \in \cX_j$. 
\eprf

We next want to prove Theorem~\ref{thm:calibration}.  We need a
preliminary lemma that shows that, in a sense, conditioning on some sets
is always at least as good as any other update rule.
\begin{lemma}
\label{lem:calwidecond}
For every probability update rule $\Pi$
and closed set $\cP \subseteq \Delta(\cX \times \cY)$, if $\Pi$ is
calibrated relative to $\cP$, then there exists a
partition $\C$ of $\cX$ such that $\C$-conditioning is narrower 
than
$\Pi$ relative to $\cP$.
\end{lemma}
\prf Suppose that $\Pi$ is calibrated relative to 
$\cP$.
Suppose that ${\bf R} = \{ {\cR_1}, \ldots, {\cR_k} \}$ (where ${\bf R}$ is as
defined in (\ref{eq:range})).  Let
$\C := \{ \cX_{\cR_1},
\ldots, \cX_{\cR_k} \}$. Then $\C$ is a partition of $\cX$.  
We want to show that $\C$-conditioning is narrower than $\Pi$ relative
to $\cP$.  Thus, we need to show that for all $x \in \cX_{\cR_j}$, 
$\cP \mid X \in \cX_{\cR_j} \subseteq \Pi(\cP,x)$.  By assumption, 
$\Pi(\cP,x) = \cR_j$.  The result is now immediate from the definition
of calibration.
\eprf

\othm{thm:calibration}
There exists an update rule that is sharply calibrated.  Moreover,
every sharply calibrated update rule is 
a generalized conditioning update rule. 
\eothm

\prf 
\commentout{
and suppose first that for some $\C_1$,
$\C_1$-conditioning is not sharply calibrated relative to $\cP$. By
Lemma~\ref{lem:condcal}, $\C_1$-conditioning must be calibrated; thus
it must be the case that the calibration is not sharp, i.e. there
exists some update rule $\Pi'$, calibrated relative to $\cP$, that is
strictly narrower than $\C_1$-conditioning relative to $\cP$. By
Lemma~\ref{lem:calwidecond}, $\Pi'$ is wider than $\C_2$-conditioning
relative to $\cP$, for some partition $\C_2$.

Now suppose, by way of contradiction, that there exists no partition
$\C$ for which $\C$-partitioning is sharply calibrated relative to
$\cP$.  Fix some arbitrary partition $\C_1$. Under our assumption
$\C_1$ is not sharply calibrated relative to $\cP$, so by the
reasoning above there exists some partition $\C_2$ such that
$\C_2$-conditioning is strictly narrower than $\C_1$-conditioning
relative to $\cP$. Since $\C_2$ is not sharply calibrated relative to
$\cP$, there also exists some partition $\C_3$ such that
$\C_3$-conditioning is a strictly narrower than both $\C_1$- and
$\C_2$-conditioning relative to $\cP$ (note that narrowing is
transitive). In general, for every positive integer $k$, there
exists a sequence of partitions $\C_1, \ldots \C_k$ such that, for $j
\in\{2, \ldots, k \}$, $\C_j$-conditioning is strictly narrower than $\C_{j-1}$-conditioning
relative to $\cP$, so
that in particular,
\begin{equation}
\label{eq:alldifferent}
\mbox{For all\ } j, j' \in \{1, \ldots, k \} \; : \; 
j \neq j' \Rightarrow \C_j \neq \C_{j'}.
\end{equation}
But then (\ref{eq:alldifferent}) must also hold for some $k$ that is
larger than the (finite) number of ways in which $\cX$ can be
partitioned.  We have arrived at the desired contradiction.
}
To show that there exists an update rule that is sharply calibrated, we
actually construct a generalized conditioning rule that is sharply
calibrated.  It suffices to show that for each closed $\cP \subseteq
\Delta(\cX \times \cY)$, 
there exists some partition $\C$ such that $\C$-conditioning is sharply
calibrated relative to $\cP$.  We can place a partial order
$\le_{\cP}$ on partitions $\C$ by taking $\C_1 \le_{\cP} \C_2$ if
$\C_1$-conditioning is narrower than $\C_2$ conditioning relative to
$\cP$.  Since $\cX$ is finite, there are only finitely many possible
partitions of of $\cX$.  Thus, there must be some minimal elements of
$\cP$.  We claim that each minimal element of $\le_{\cP}$ is sharply
calibrated relative to $\cP$.  For suppose that $\C$ is minimal relative
to $\le_{\cP}$.  If $\Pi$ is an update rule that is strictly narrower
than $\C$ relative to $\cP$, then, by Lemma~\ref{lem:calwidecond}, there
exists a partition $\C'$ such that $\C'$ is narrower than $\Pi$ relative
to $\cP$.  But then $\C' <_{\cP} \C$, contradicting the minimality of
$\C$.  This proves the desired result.

To show that every sharply calibrated update rule is a generalized
conditioning rule, suppose that
$\Pi$ is sharply calibrated.  Given $\cP$, by
Lemma~\ref{lem:calwidecond}, 
there must be some partition $\C$ such that $\C$-conditioning is
narrower that $\Pi$, relative to $\cP$.
By Proposition~\ref{pro:calibration},
$\C$-conditioning is calibrated relative to $\cP$.
Since $\Pi$ is {\em sharply\/} calibrated, there can be no
$\Pi'$ that is strictly narrower than $\Pi$ relative to $\cP$ and that
is also calibrated relative to $\cP$. 
Thus, $\Pi$ must in fact coincide with
$\C$-conditioning  relative to $\cP$. 
This proves that $\Pi$ is a generalized conditioning rule.
\eprf

\commentout{
\label{app:dawidproofs}
\subsection{Proofs of Section~\ref{sec:maxent}}
We now prove Theorem~\ref{thm:dawida} and~\ref{thm:dawidb}, both of
which refer to Nash equilibria involving decision rules $\delta: \cX
\rightarrow \rand$ and the space $\cX \times \cY$. This is
done by first stating and proving Theorem~\ref{thm:dawidc}
and~\ref{thm:dawidd}. These are 
two completely analogous theorems involving just
actions $\rand$ and the space $\cY$. 
\begin{theorem}
\label{thm:dawidc}
Let $L$ be simple, and let 
$\cY$, $\cPy \subseteq \Delta(\cY)$ be arbitrary.
\begin{itemize}
\item[a.] Suppose that there exists $\Py^* \in \cPy$ and $a^* \in
  \rand$ such that $(\Py^*, a^*)$ is a Nash equilibrium in the
  game defined by $L$ and $\cP$, i.e. for all $a \in \rand$,
  $E_{\Py^*}[L(Y,a)] \geq E_{\Py^*}[L(Y,a^*)]$, and for all $\Py 
\in
  \cPy$, $E_{\Py}[L(Y,a^*)] \leq E_{\Py^*}[L(Y,a^*)]$. Then $\Py^*$ is
  a maximin strategy and $a^*$ is a minimax act, i.e.
$$
\begin{array}{ccl}
E_{\Py^*} [L(Y,a^*)] & = & 
{\min}_{a \in \rand} 
E_{\Py^*} [L(Y,a)]  = 
{\max}_{\Py \in \cPy} {\min}_{a \in \rand} 
E_{\Py} [L(Y,a)] \\
& = &  
{\min}_{a\in \rand} 
{\max}_{\Py \in \cPy} E_{\Py} [L(Y,a)] = 
{\max}_{\Py \in \cPy} E_{\Py} [L(Y,a^*)].
\end{array}
$$ 
Note in particular that $a^*$ is a Bayes act relative to $\Py^*$.
\item[b.]
Now let $\cY$ be finite and let $\cPy \subseteq \Delta(\cY)$ be closed and
convex. Then a Nash equilibrium $(\Py^*, a^*)$ with finite value $E_{\Py^*}[L(Y,a^*)]$ exists in the game.
\end{itemize}
\end{theorem}
\prf
Part a. is just Theorem 4.1 of \cite{GrunwaldD04} (replace the phrase
`saddle point' by `Nash equilibrium').
If $\cA$ is finite and $L$ is bounded, then Part b. is just Von
Neumann's \citeyear{VonNeumann28} original minimax theorem.
Otherwise, using the fact that $L$ is finite on the interior of
$\rand$, denoted by $\interior(\rand)$, 
we can apply 
Theorem 6.1 of \cite{GrunwaldD04} to the game with actions restricted to the interior of $\rand$, which gives (a) that
\begin{equation}
\label{eq:ferguson}
{\sup}_{\Py \in \cPy} {\inf}_{a \in \interior(\rand)} 
E_{\Py}[L(Y,a)] = 
{\inf}_{a \in \interior(\rand)} {\sup}_{\Py \in \cPy} E_{\Py}[L(Y,a)],
\end{equation}
and (b) that the supremum 
on the left is achieved by some $\Py^* \in \cPy$.
By continuity of $L$ and compactness of $\rand$, ${\inf}_{a \in \interior(\rand)}
E_{\Py}[L(Y,a)] = {\inf}_{a \in \rand} E_{\Py}[L(Y,a)]$.
By Proposition~\ref{pro:mmactexists}, for each $\Py$, the infimum is achieved, so that we have
\begin{equation}
\label{eq:supinf}
{\sup}_{\Py \in \cPy} {\inf}_{a \in \interior(\rand)} 
E_{\Py}[L(Y,a)] = {\max}_{\Py \in \cPy} {\min}_{a \in \rand} 
E_{\Py}[L(Y,a)].
\end{equation}
On the other hand, suppose for some $a' \in \rand \setminus
\interior(\rand)$, it holds that ${\sup}_{\Py \in \cPy} E_{\cP}[L(Y,a')]
\leq {\sup}_{\Py \in \cPy} {\inf}_{a \in \interior(\rand)}
E_{\cP}[L(Y,a)]$. Then $L(y,a') < \infty$ for all $y$ in the support of
$\cPy$. Therefore, by continuity of $L$, for all $\epsilon > 0$ there exists 
$a \in \interior(\rand)$ with 
$$|{\sup}_{\Py \in \cPy} E_{\Py}[L(Y,a')] -
{\sup}_{\cPy} 
E_{\Py}[L(Y,a)]| \leq {\max}_{y \in \cY} |L(y,a') - L(y,a)| <   \epsilon,
$$ 
which gives  ${\sup}_{\Py \in \cPy} E_{\cP}[L(Y,a')]
= {\sup}_{\Py \in \cPy} {\inf}_{a \in \interior(\rand)}
E_{\cP}[L(Y,a)]$. Together with Proposition~\ref{pro:mmactexists}, this implies that 
\begin{equation}
\label{eq:infsup}
{\inf}_{a \in \interior(\rand)}  {\sup}_{\Py \in \cPy} 
E_{\Py}[L(Y,a)] = {\min}_{a \in \rand}  {\max}_{\Py \in \cPy} 
E_{\Py}[L(Y,a)].
\end{equation}
The result now follows by combining (\ref{eq:ferguson}),
(\ref{eq:supinf}) and (\ref{eq:infsup}), and noting that we already
proved above that the maximum on the left is achieved by some $\Py^* \in
\cPy$, Proposition~\ref{pro:mmactexists} implies that 
the minimum on the right is achieved by some  $a^* \in \rand$, and that the expression in (\ref{eq:ferguson}) is finite.  Then, by Lemma 4.1 of \cite{GrunwaldD04},
$(\Py^*,a^*)$ must be a Nash equilibrium of the game.  \eprf
\begin{theorem}
\label{thm:dawidd}
Let $\cY$ be finite, let $L$ be simple and let $\cPy \subset
\Delta(\cY)$ be arbitrary. Then
\begin{itemize}
\item[a.]
$H_L(\Py)$ is a concave, continuous and bounded 
function on $\Delta(\cY)$.
If $L$ is strictly proper then $H_L(\Py)$ is strictly concave.

\item[b.] If $(\Py^*,a^*)$ is a Nash equilibrium of the game defined by
  $L$ and $\cP$,
then $\Py^*$ must be a {\em generalized maximum entropy distribution},
achieving $\max_{\Py \in \cPy} H_L(\Py)$. 
\item[c.] If $\cP$ is closed and convex and 
$L$ is strictly proper then the maximum is achieved by a unique $P^*
\in \cPy$, and then $a^*$ is the unique minimax act 
and $(\Py^*,a^*)$ is the unique Nash equilibrium of the game.
\end{itemize}
\end{theorem}
\prf
We first prove Part a. By Proposition 3.2 of \cite{GrunwaldD04},
$H_L(\Py)$ is concave on $\Delta(\cY)$, and hence continuous on the
interior of $\Delta(\cY)$. Since we assume $L$ bounded from below,
$H(\Py)$ is bounded from below. If $L$ is bounded from above, then
$H(P)$ is also bounded from above and hence finite. If $L$ is not bounded from above,
then by our continuity condition on $L$, there exists an act $a$ such
that $E_{\Py}[L(Y,a)]$ is finite. Since $H_L(\Py) \leq E_{\Py}[L(Y,a)]$, it
is also finite. Thus, for all simple
$L$,  $H(\Py)$ is finite for all $\Py \in
\Delta(\cY)$. It then follows from Corollary 3.3. of
\cite{GrunwaldD04} that $H_L(\Py)$ is continuous on the whole of $\Delta(\cY)$.

Now suppose $L$ is strictly proper. Let $\P_0, \P_1 \in \Delta(\cY)$ and
define $P_{\lambda} = \lambda P_1 + (1- \lambda) P_0$. Then for $0 <
\lambda < 1$,
$$
\begin{array}{rcl}
\lambda H(P_1) + (1- \lambda) H(P_0) & = &
\lambda E_{P_1}[L(Y,P_1)] + (1- \lambda) E_{P_0}[L(Y,P_0)]  \\
& < & 
\lambda E_{P_1}[L(Y,P_\lambda)] + (1- \lambda) E_{P_0}[L(Y,P_\lambda)]
\\
& = & E_{P_{\lambda}}[L(Y,P_{\lambda})] = H(\lambda P_1 + (1- \lambda) P_0),
\end{array}
$$
where the inequality follows by strict properness. This shows that $H$
is strictly concave on $\Delta(\cY)$.

Part b. follows from the definition
of $H_L$. 

For Part c., note that, since, if $L$ is strictly proper, $H_L(P)$ is strictly
concave, and since $\cP$ is convex, it follows that $P^*$ uniquely achieves
$\max_{P \in \cP} H_L(P)$. Since, again by strict properness, $P^*$
has a unique Bayes act equal to $a^*$, it further follows that $a^*$
is the unique a priori minimax decision rule, so that, by
Theorem~\ref{thm:dawida},  $(P^*,a^*)$ must
be the unique Nash equilibrium of the game. 
\eprf

The proofs of  Theorem~\ref{thm:dawida} and~\ref{thm:dawidb} of the main
text now proceed by defining a new loss function $L'$ that allows us to
connect them to the two theorems above.
To this end, define $\cY' = \cX \times \cY$. Denote the elements of
$\cX$ as $\{1, \ldots, k \}$. Define
$\cA'$ as the set of all vectors $(\delta(1), \ldots, \delta(k))$
satisfying, for all $x \in \cX$, $\delta(x) \in \rand$. Finally,  
define the loss function
$L': \cY' \times \cA' \rightarrow \reals \cup [\infty]$ as 
$$
L'((x,y),\delta) := L(y, \delta(x)).
$$
\pro 
\label{prop:reduction}
$L'$ is a simple loss function relative to $\cY'$ and $\cA'$.  \epro
\prf If $\cA$ is finite, then $\cA'$ is a closed and convex subset of
$\reals^{|\cX | \times |\cA|}$ with nonempty interior, and $L'$
satisfies the continuity condition of Definition~\ref{def:simple}.
Therefore $L'$ is simple.  If $\cA$ is infinite, then $\cA'$ is a
closed and convex subset of $\reals^{|\cX | \times m}$ for the $m$
with $\cA \subseteq \reals^m$; again $\cA'$ has nonempty interior, and
$L'$ satisfies the continuity condition of Definition~\ref{def:simple}. Therefore $L'$ is
simple. \eprf

We now have all the necessary tools to prove Theorem~\ref{thm:dawida}
and~\ref{thm:dawidb}.

\prf {\bf (of Theorem~\ref{thm:dawida})}
By Proposition~\ref{prop:reduction} and the fact that 
for all $P \in \Delta(\cX \times \cY)$, $H^X_L(P) = H_{L'}(P)$,
Theorem~\ref{thm:dawida} is just a special case of Theorem~\ref{thm:dawidc}.
\eprf

\prf {\bf (of Theorem~\ref{thm:dawidb})}
(Part a) By Proposition~\ref{prop:reduction} and the fact that 
for all $P \in \Delta(\cX \times \cY)$, $H^X_L(P) = H_{L'}(P)$,
the fact that $H^X_L(P)$ is continuous, concave and
bounded follows from the corresponding statement of Theorem~\ref{thm:dawidd}.

Now if $L$ is strictly proper and $\Px \in \Delta(\cX)$, then the set $\cP' = \{ \Pr \in \Delta(\cX \times \cY) : \Pr_{\cX} = \Px \}$ is convex.  Denoting $\cX = \{1, \ldots, k \}$, we can write, for all $\Pr \in \cP'$, 
$$
H^X_L(\Pr) = \sum_{x=1}^k \alpha_x H_L((\Pr \mid X=x)_{\cY})
$$
for some $\alpha_1, \ldots, \alpha_k$. By Theorem~\ref{thm:dawidd},
$H_L$ is strictly concave on $\Delta(\cY)$. Since $\Pr(Y=y \mid X=x)=
\Pr(x,y)/\alpha_x$ and $\alpha_x$ is fixed on $\cP'$, we have, for $x = 1..k$,
$H_L((\Pr \mid X= x)_{\cY})$ is strictly concave on $\cP'$. Therefore,
$H_L^X(\Pr)$ is a mixture of strictly concave functions on a convex
set, and must thus itself be strictly concave.

(Part b) 
By Proposition~\ref{prop:reduction} and the fact that 
for all $\Pr \in \Delta(\cX \times \cY)$, $H^X_L(\Pr) = H_{L'}(\Pr)$,
the fact that $\Pr^*$ must achieve  $\max_{\Pr \in \cP} H^X_L(\Pr)$ follows from the corresponding statement of Theorem~\ref{thm:dawidd}.

(Part c) Suppose that $H^* := \max_{\Pr \in \cP} H^X_L(\Pr)$ is
achieved for two distributions $\Pr_1$ and $\Pr_0 \neq \Pr_1$. Let
$\Pr_{\lambda} = \lambda \Pr_1 + (1- \lambda) \Pr_0$. Then for all $0
\leq \lambda \leq 1$, $\Pr_\lambda \in \cP$, and, since we already
established that $H_L^X(\Pr)$ is concave on $\cP$, we must have
$H^X_L(\Pr_{\lambda}) = H^*$. Since $L$ is strictly proper,
$\delta_{\lambda}$, the Bayes decision rule for ${\Pr}_{\lambda}$, satisfies
$\delta_{\lambda}(x) = ({\Pr}_{\lambda} | X=x)_{\cY}$. In particular,
$$
\sum_{x,y} {{\Pr}}_{0.5}(x,y) L(y,\delta_{0.5}(x)) = H_L^X({\Pr}_{0.5})
= H_L^X({\Pr}_0) 
= H_L^X({\Pr}_1). 
$$
But then, since $E_{\Pr_{\lambda}} [L(Y, \delta_{0.5}(X))]$ is
linear as a function of $\Pr_{\lambda}$ and coincides with
$H_L^X({\Pr}_{0.5})$ at $\lambda = 0.5$, we must either have $E_{{\Pr}_0}[L(Y,\delta_{0.5}(X)] < H^X_L({\Pr}_0)$ or $E_{{\Pr}_1}[L(Y,\delta_{0.5}(X)] < H^X_L({\Pr}_1)$ or
$$E_{{\Pr}_0}[L(Y,\delta_{0.5}(X)] = H_L^X({\Pr}_{0.5}) =
E_{{\Pr}_1}[L(Y,\delta_{0.5}(X)].$$
The first two possibilities
contradict the definition of $H^X_L$ so the third must be the case.
But since $L$ is strictly proper, the third possibility implies that
${\Pr}_{0.5} \mid X= x = {\Pr}_1 \mid X= x$ for all $x$ with ${\Pr}_0(X=x) > 0$ or
${\Pr}_1(X=x) > 0$. Since ${\Pr}_0$ and ${\Pr}_1$ were arbitrary elements of
the support of $\cP^*$, the result about ${\Pr}^* \mid X=x$ follows. The
result about $\delta^*$ then follows by strict properness.  \eprf

\subsection{Proofs of Section~\ref{sec:aprioriconditioning}}
\prf {\bf (of Lemma~\ref{lem:ignoremm})}
Let $a^*$ be minimax optimal against $\cPy_{\cY}$. If
$\max_{{\Pr} \in \cP} H_L^X({\Pr}) = 
\max_{\Py \in \cP_{\cY}} H_L(P)$, then
\begin{equation}
\label{eq:wolf}
\begin{array}{rcl}
{\max}_{{\Pr} \in \cP} E_{{\Pr}}[L_{\dprior}] & = &
 {\max}_{\Py \in \cP_{\cY}} E_{\Py}[L(Y,a^*)] \\
 & = & {\max}_{{\Pr} \in \cP} E_{{\Pr}} [L(Y,a^*)]\\
& = & {\max}_{{\Pr} \in \cP} E_{{\Pr}} [L_{\dign}],
\end{array}
\end{equation}
so that $\dign$ is an a priori minimax decision rule.

For the converse, first note that, by concavity of $H_L$ and Jensen's
inequality,
$$
{\max}_{{\Pr} \in \cP} H_L^X({\Pr}) = 
{\max}_{{\Pr} \in \cP} \sum_{x \in \cX} {\Pr}(X=x) H_L(({\Pr} \mid X=x)_{\cY}) \leq
{\max}_{\Py \in \cP_{\cY}} H_L(\Py).
$$
Repeating (\ref{eq:wolf}) with the equality replaced by a
strict inequality, if ${\max}_{{\Pr} \in \cP} H_L^X({\Pr}) < {\max}_{\Py \in
  \cP_{\cY}} H_L(\Py)$, then $\max_{{\Pr} \in \cP} E_{{\Pr}}[L_{\dprior}] <
\max_{{\Pr} \in \cP} E_{{\Pr}} [L_{\dign}]$, so that $\dign$ is not an a
priori minimax decision rule.  \eprf

\ \\
The proof of Theorem~\ref{thm:ignoremm} is based on
Proposition~\ref{pro:expose} below, which we state and prove first. The
proposition is based on a particular loss function, the so-called {\em Brier
  score relative to distribution $\Qy$} \cite{GrunwaldD04}. While the
definition of the relative Brier score may look unfamiliar, the
corresponding generalized entropy of $\Py$ is the familiar $\ell_2$-distance
between $\Py$ and $\Qy$.

Formally, let $\cA = \Delta(\cY)$ and let 
$\Qy \in \Delta(\cY)$. Following Section 8.5.1 of \citeN{GrunwaldD04}, the Brier score $L_{\Qy}$ relative to $\Qy$ is defined as 
$$
L_{\Qy}(y,\Py) := \sum_{j \in \cY} (I_{Y=j} - \Py(y))^2 - \sum_{j
  \in \cY} (I_{Y=j} - \Qy(y))^2,
$$
where $I_{Y=j}$ is the indicator random variable, which is $1$ if
$Y=j$ and $0$ otherwise. The following proposition follows by straightforward rewriting of this definition.
\pro
\label{pro:expose}
For each $\Qy, \Py \in \Delta(\cY)$, $\min_{R \in \Delta(\cY)}
E_{\Py}[L_{\Qy}(Y,R)]$ is uniquely achieved for $R = \Py$. Therefore,
\begin{itemize}
\item[a.]
The relative Brier score is a strictly proper loss function.
\item[b.] $H_{L_{\Qy}}(\Py) = E_{\Py}[L_{\Qy}(Y,\Py)] = - \sum_{y \in
    \cY} (\Py(y) - \Qy(y))^2$, so that, if $\cPy$ is a set of distributions on $\cY$ 
containing $\Qy$, then $\max_{\Py \in \cPy} H_{L_{\Qy}}(\Py)$ is uniquely achieved for $\Py = \Qy$.
\end{itemize}
\epro

For all $\cP \subseteq \Delta(\cX \times \cY)$ we have: (a) if $\Pi$ is sharply calibrated relative to $\cP$, then it is
equivalent to $\C$-conditioning relative to $\cP$, for some partition
$\C$ of $\cX$; (b)
conversely, 
there exist a partition $\C$ of $\cX$ such that
$\C$-conditioning is sharply calibrated relative to $\cP$.
\eothm
\prf 
(Part a.) Let $L$ be any simple loss function. We have, by Lemma~\ref{lem:ignoremm},
\begin{equation}
\label{eq:late}
{\max}_{{\Pr} \in \cP} H_L^X({\Pr}) \leq {\max}_{\Py \in \cP_{\cY}} H_L(\Py).
\end{equation}
By the condition of the theorem, for each $\Py \in \cP_{\cY}$, $\cP$
contains a distribution ${\Pr}$ such that $H^X_L({\Pr}) = H_L(\Py)$. It
follows that (\ref{eq:late}) holds with equality, whence again by
Lemma~\ref{lem:ignoremm}, $\dign$ is a priori minimax optimal.  

(Part b.) By Proposition~\ref{pro:expose}, there exists a simple,
strictly proper loss function $L$ such that $\max_{\Py \in \cP_{\cY}}
H_L(\Py)$ is uniquely achieved for the $\Py^*$ mentioned in the
theorem.  Now note $H_L^X({\Pr}) = \sum_{x \in \cX} {\Pr}(X=x) H_L(({\Pr} \mid
X=x)_{\cY})$. Therefore, for each ${\Pr} \in \cP$, $H_L^X({\Pr}) < H_L(\Py^*)$. Since $\cP$
is compact and convex and $H_L$ is continuous, it follows that
$\max_{{\Pr} \in \cP} H_L^X({\Pr}) < H_L(\Py^*)$. It now follows from
Lemma~\ref{lem:ignoremm} that, relative to $L$, $\dign$ is not a
priori minimax optimal. Here $L$ is a strictly proper loss function.

We can transform the result to a loss function with finite $\cA$ by
discretization, as follows. For all $\epsilon > 0$, by discretizing
$\cA = \cP$ to sufficient precision, we can construct a loss function
$L'$ relative to finite set $\cA' \subseteq \cA$ such that for all $y
\in \cY$, $a \in \cA$, $|L(y,a) - L'(y,a')| < \epsilon$, where $a'$ is
the closest point in $\cA'$ to $a \in \cA$. It now follows by
continuity of $H$ and compactness of $\cP$ that, for small enough
$\epsilon$, the corresponding $L'$ will be such that $\max_{{\Pr} \in
  \cP} H_{L'}^X({\Pr}) < \max_{\Py \in \cP_{\cY}} H_{L'}(\Py^*)$, so that
once again $\dign$ is not a priori minimax optimal relative to $L'$ and $\cA'$.
\eprf
}
\end{document}

%% file: defn.tex
\newtheorem{THEOREM}{Theorem}[section]
\newenvironment{theorem}{\begin{THEOREM} \hspace{-.85em} {\bf :} }%
                        {\end{THEOREM}}
\newtheorem{LEMMA}[THEOREM]{Lemma}
\newenvironment{lemma}{\begin{LEMMA} \hspace{-.85em} {\bf :} }%
                      {\end{LEMMA}}
\newtheorem{COROLLARY}[THEOREM]{Corollary}
\newenvironment{corollary}{\begin{COROLLARY} \hspace{-.85em} {\bf :} }%
                          {\end{COROLLARY}}
\newtheorem{PROPOSITION}[THEOREM]{Proposition}
\newenvironment{proposition}{\begin{PROPOSITION} \hspace{-.85em} {\bf :} }%
                            {\end{PROPOSITION}}
\newtheorem{DEFINITION}[THEOREM]{Definition}
\newenvironment{definition}{\begin{DEFINITION} \hspace{-.85em} {\bf :} \rm}%
                            {\end{DEFINITION}}
\newtheorem{CLAIM}[THEOREM]{Claim}
\newenvironment{claim}{\begin{CLAIM} \hspace{-.85em} {\bf :} \rm}%
                            {\end{CLAIM}}
\newtheorem{EXAMPLE}[THEOREM]{Example}
\newenvironment{example}{\begin{EXAMPLE} \hspace{-.85em} {\bf :} \rm}%
                            {\end{EXAMPLE}}
\newtheorem{REMARK}[THEOREM]{Remark}
\newenvironment{remark}{\begin{REMARK} \hspace{-.85em} {\bf :} \rm}%
                            {\end{REMARK}}

\newcommand{\thm}{\begin{theorem}}
\newcommand{\lem}{\begin{lemma}}
\newcommand{\pro}{\begin{proposition}}
\newcommand{\dfn}{\begin{definition}}
\newcommand{\rem}{\begin{remark}}
\newcommand{\xam}{\begin{example}}
\newcommand{\cor}{\begin{corollary}}
\newcommand{\prf}{\noindent{\bf Proof:} }
\newcommand{\ethm}{\end{theorem}}
\newcommand{\elem}{\end{lemma}}
\newcommand{\epro}{\end{proposition}}
\newcommand{\edfn}{\bbox\end{definition}}
\newcommand{\erem}{\bbox\end{remark}}
\newcommand{\exam}{\bbox\end{example}}
\newcommand{\ecor}{\end{corollary}}
\newcommand{\eprf}{\bbox\vspace{0.1in}}
\newcommand{\beqn}{\begin{equation}}
\newcommand{\eeqn}{\end{equation}}

\newcommand{\bbox}{\vrule height7pt width4pt depth1pt}

\newcommand{\clm}{\begin{claim}}
\newcommand{\eclm}{\end{claim}}

\newcommand{\union}{\cup}
\newcommand{\inter}{\cap}

\newcommand{\IR}{\mbox{$I\!\!R$}}

\newcommand{\IN}{\mbox{$I\!\!N$}}

\renewcommand{\phi}{\varphi}

\newcommand{\A}{{\cal A}}
\newcommand{\B}{{\cal B}}
\newcommand{\C}{{\cal C}}

\renewcommand{\P}{{\cal P}}

\newcommand{\X}{{\cal X}}
\newcommand{\Y}{{\cal Y}}

\newcommand{\ol}{\setlength{\itemsep}{0pt}\begin{enumerate}}
\newcommand{\eol}{\end{enumerate}\setlength{\itemsep}{-\parsep}}
\newcommand{\ul}{\setlength{\itemsep}{0pt}\begin{itemize}}
\newcommand{\dl}{\setlength{\itemsep}{0pt}\begin{description}}
\newcommand{\edl}{\end{description}\setlength{\itemsep}{-\parsep}}
\newcommand{\eul}{\end{itemize}\setlength{\itemsep}{-\parsep}}

\newcommand{\cA}{{\cal A}}

\newcommand{\cR}{{\cal R}}
\newcommand{\cS}{{\cal S}}

\newcommand{\commentout}[1]{}

\newcommand{\bi}{\begin{itemize}}
\newcommand{\ei}{\end{itemize}}
\newcommand{\be}{\begin{enumerate}}
\newcommand{\ee}{\end{enumerate}}

%% file: usopcorr.bbl
\begin{thebibliography}{}

\bibitem[\protect\citeauthoryear{Augustin}{Augustin}{2003}]{Augustin03}
Augustin, T. (2003).
\newblock On the suboptimality of the generalized {B}ayes rule and robust
  {B}ayesian procedures from the decision theoretic point of view:~{A}
  cautionary note on updating imprecise priors.
\newblock In {\em 3rd International Symposium on Imprecise Probabilities and
  Their Applications}, pp.\  31--45.
\newblock Available at http://www.carleton-scientific.com/isipta/2003-toc.html.

\bibitem[\protect\citeauthoryear{Cozman and Walley}{Cozman and
  Walley}{2001}]{CozmanWalley}
Cozman, F.~G. and P.~Walley (2001).
\newblock Graphoid properties of epistemic irrelevance and independence.
\newblock In {\em 2nd International~Symposium on~Imprecise Probabilities and
  Their Applications}, pp.\  112--121.
\newblock Available at http://www.sipta.org/~isipta01/proceedings/index.html.

\bibitem[\protect\citeauthoryear{Dawid}{Dawid}{1982}]{Dawid82}
Dawid, A. (1982).
\newblock The well-calibrated {Bayesian}.
\newblock {\em Journal of the American Statistical Association\/}~{\em 77},
  605--611.
\newblock Discussion: pages 611--613.

\bibitem[\protect\citeauthoryear{G\"ardenfors and Sahlin}{G\"ardenfors and
  Sahlin}{1982}]{GS82}
G\"ardenfors, P. and N.~Sahlin (1982).
\newblock Unreliable probabilities, risk taking, and decision making.
\newblock {\em Synthese\/}~{\em 53}, 361--386.

\bibitem[\protect\citeauthoryear{Gilboa and Schmeidler}{Gilboa and
  Schmeidler}{1989}]{GS1989}
Gilboa, I. and D.~Schmeidler (1989).
\newblock Maxmin expected utility with a non-unique prior.
\newblock {\em Journal of Mathematical Economics\/}~{\em 18}, 141--153.

\bibitem[\protect\citeauthoryear{Grove and Halpern}{Grove and
  Halpern}{1998}]{GroveHalpern98}
Grove, A.~J. and J.~Y. Halpern (1998).
\newblock Updating sets of probabilities.
\newblock In {\em Proc.~Fourteenth Conference on Uncertainty in Artificial
  Intelligence (UAI '98)}, pp.\  173--182.

\bibitem[\protect\citeauthoryear{Gr\"unwald and Dawid}{Gr\"unwald and
  Dawid}{2004}]{GrunwaldD04}
Gr\"unwald, P. and A.~Dawid (2004).
\newblock Game theory, maximum entropy, minimum discrepancy, and robust
  {B}ayesian decision theory.
\newblock {\em The Annals of Statistics\/}~{\em 32\/}(4), 1367--1433.

\bibitem[\protect\citeauthoryear{Gr\"unwald and Halpern}{Gr\"unwald and
  Halpern}{2004}]{GrunwaldH04}
Gr\"unwald, P. and J.~Halpern (2004).
\newblock When ignorance is bliss.
\newblock In {\em Proc.~Twentieth Conference on Uncertainty in Artificial
  Intelligence (UAI 2004)}, pp.\  226--234.

\bibitem[\protect\citeauthoryear{Herron, Seidenfeld, and Wasserman}{Herron
  et~al.}{1997}]{HerronSW97}
Herron, T., T.~Seidenfeld, and L.~Wasserman (1997).
\newblock Divisive conditioning: Further results on dilation.
\newblock {\em Philosophy of Science\/}~{\em 64}, 411--444.

\bibitem[\protect\citeauthoryear{Hughes and {van Fraassen}}{Hughes and {van
  Fraassen}}{1985}]{vF3}
Hughes, R. I.~G. and B.~C. {van Fraassen} (1985).
\newblock Symmetry arguments in probability kinematics.
\newblock In P.~Kitcher and P.~Asquith (Eds.), {\em PSA 1984}, Volume~2, pp.\
  851--869. East Lansing, Michigan: Philosophy of Science Association.

\bibitem[\protect\citeauthoryear{Mosteller}{Mosteller}{1965}]{Mosteller}
Mosteller, F. (1965).
\newblock {\em Fifty Challenging Problems in Probability with Solutions}.
\newblock Reading, Mass.: Addison-Wesley.

\bibitem[\protect\citeauthoryear{Seidenfeld}{Seidenfeld}{2004}]{Seidenfeld04}
Seidenfeld, T. (2004).
\newblock A contrast between two decision rules for use with (convex) sets of
  probabilities: $\gamma$-maximin versus {$E$}-admissibility.
\newblock {\em Synthese\/}.
\newblock To appear.

\bibitem[\protect\citeauthoryear{Seidenfeld and Wasserman}{Seidenfeld and
  Wasserman}{1993}]{SeidenfeldW93}
Seidenfeld, T. and L.~Wasserman (1993).
\newblock Dilation for convex sets of probabilities.
\newblock {\em Annals of Statistics\/}~{\em 21}, 1139--1154.

\bibitem[\protect\citeauthoryear{{van Fraassen}}{{van Fraassen}}{1987}]{vF1}
{van Fraassen}, B.~C. (1987).
\newblock Symmetries of personal probability kinematics.
\newblock In N.~Rescher (Ed.), {\em Scientific Enquiry in Philsophical
  Perspective}, pp.\  183--223. Lanham, Md.: University Press of America.

\bibitem[\protect\citeauthoryear{{vos Savant}}{{vos Savant}}{1990}]{vScomb}
{vos Savant}, M. (Sept. 9, 1990).
\newblock Ask {M}arilyn.
\newblock {\em Parade Magazine\/}, 15.
\newblock Follow-up articles appeared in {\em Parade Magazine} on Dec.~2, 1990
  (p.~25) and Feb.~17, 1991 (p. 12).

\bibitem[\protect\citeauthoryear{Vovk, Gammerman, and Shafer}{Vovk
  et~al.}{2005}]{VovkGS05}
Vovk, V., A.~Gammerman, and G.~Shafer (2005).
\newblock {\em Algorithmic Learning in a Random World}.
\newblock New York: Springer.

\bibitem[\protect\citeauthoryear{Wald}{Wald}{1950}]{Wald50}
Wald, A. (1950).
\newblock {\em Statistical Decision Functions}.
\newblock New York: Wiley.

\bibitem[\protect\citeauthoryear{Walley}{Walley}{1991}]{Walley91}
Walley, P. (1991).
\newblock {\em Statistical Reasoning with Imprecise Probabilities}, Volume~42
  of {\em Monographs on Statistics and Applied Probability}.
\newblock London: Chapman and Hall.

\end{thebibliography}
